\documentclass[10pt,letterpaper,twocolumn]{article}

\PassOptionsToPackage{usenames,dvipsnames}{xcolor}

\usepackage{sty/usenix-2020-09}

\PassOptionsToPackage{breaklinks,colorlinks}{hyperref}
\usepackage[breaklinks,colorlinks]{hyperref}
\hypersetup{
  colorlinks,
  linkcolor={red!50!black},
  citecolor={blue!50!black},
  urlcolor={blue!50!black}
}

\usepackage{amsmath,amsopn}
\usepackage{cleveref}%needs to be loaded *after* hyperref and amsmath
\crefname{section}{\S}{\SS}
\crefformat{section}{\S#2#1#3} % see manual of cleveref, section 8.2.1
\crefformat{subsection}{\S#2#1#3}
\crefformat{subsubsection}{\S#2#1#3}
\usepackage{endnotes,microtype,xspace,fancyvrb,multirow}
\usepackage{tikz}
\usepackage{booktabs} % comment out if  ACM
\usepackage{array,underscore,relsize}
\usepackage[T1]{fontenc}
\usepackage{enumitem}
\usepackage[belowskip=0pt,aboveskip=2pt,labelfont=bf]{subcaption}
\newcommand{\sys}{\mbox{\textsc{ReInc}}\xspace}

\fvset{fontsize=\scriptsize,xleftmargin=8pt,numbers=left,numbersep=5pt}

\input{code/fmt}

\newcommand{\x}{$\times$\xspace}

\newif\ifdraft\drafttrue
\newif\ifnotes\notestrue
\ifdraft\else\notesfalse\fi

\input{glyphtounicode}
\newcolumntype{R}[1]{>{\raggedleft\let\newline\\\arraybackslash\hspace{0pt}}p{#1}}

\newcommand{\squishlist}{
\begin{itemize}[noitemsep,nolistsep]
  \setlength{\itemsep}{-0pt}
}
\newcommand{\squishend}{
  \end{itemize}
}

\DeclareRobustCommand\circledtikz[1]{\tikz[baseline=(char.base)]{
    \node[shape=circle,draw,inner sep=1pt] (char)
    {#1};}}

\DeclareRobustCommand\filledcircledtikz[1]{\tikz[baseline=(char.base)]{
    \node[shape=circle,draw,fill,inner sep=1pt] (char)
    {\textcolor{white}{#1}};}}

\usepackage{xstring}
\newcommand{\PP}[1]{
\vspace{2px}
\noindent{\bf \IfEndWith{#1}{.}{#1}{#1.}}
}

\newcommand{\ie}{{i}.{e}.}

\newcommand{\paraf}[1]{\noindent\textbf{#1}}

\usepackage{color}
\definecolor{deepblue}{rgb}{0,0,0.5}
\definecolor{deepred}{rgb}{0.6,0,0}
\definecolor{deepgreen}{rgb}{0,0.5,0}

\usepackage{listings}
\lstdefinelanguage{paper}{
 keywords={partition, transform, gather, scatter, apply},
 keywordstyle=\color{blue}\bfseries,
 morekeywords={[2]degrees,branch,commit,v_prev},
 keywordstyle={[2]\color{red}\bfseries},
 morekeywords={[3]if,def,Class,return,else,None,False,True,Array,while,G},
 keywordstyle={[3]\bfseries},
 morekeywords={[4]update, msg_pass, cache, integrate, stack_seq_model},
 keywordstyle={[4]\color{deepblue}\bfseries},
 morekeywords={[5]msg_udf, reduce_udf},
 keywordstyle={[5]\color{deepred}\bfseries},
 basicstyle=\small\ttfamily,
 identifierstyle=\color{black},
 sensitive=false,
 comment=[l]{\/\/},
 morecomment=[s]{/*}{*/},
 commentstyle=\color{green}\ttfamily,
 stringstyle=\color{red}\ttfamily,
 breaklines=true,
}

\lstnewenvironment{python}[2]
{\lstset{
    language=paper,
    frame=tlbr,
    caption= #1,
    label=#2,
    columns=fullflexible,
    postbreak=\mbox{\textcolor{red}{$\hookrightarrow$}\space},
    captionpos=b}}
{}

\usepackage[font=small, textfont=bf, labelfont=bf, aboveskip=5pt]{caption}
\usepackage{subcaption}

\usepackage{titlesec}
\titlespacing*{\section}
{0pt}{2ex}{1.5ex}
\titlespacing*{\subsection}
{0pt}{2ex}{1.5ex}
\titlespacing*{\subsubsection}
{0pt}{2ex}{1.5ex}

\usepackage{titling}%used for removing unnecessary space from \maketitle
\begin{document}

\date{}
\title{\sys: Scaling Training of Dynamic Graph Neural Networks}

\author{
{\rm Mingyu Guan}\\
Georgia Institute of Technology
\and
{\rm Saumia Singhal}\\
Georgia Institute of Technology
\and
{\rm Taesoo Kim}\\
Georgia Institute of Technology
\and
{\rm Anand Padmanabha Iyer}\\
Georgia Institute of Technology
} % end author

\maketitle

\begin{abstract}

Dynamic Graph Neural Networks (DGNNs) have gained widespread attention due to their applicability in diverse domains such as traffic network prediction, epidemiological forecasting, and social network analysis. In this paper, we present \sys, a system designed to enable efficient and scalable training of DGNNs on large-scale graphs. \sys introduces key innovations that capitalize on the unique combination of Graph Neural Networks (GNNs) and Recurrent Neural Networks (RNNs) inherent in DGNNs. By reusing intermediate results and incrementally computing aggregations across consecutive graph snapshots, \sys significantly enhances computational efficiency. To support these optimizations, \sys incorporates a novel two-level caching mechanism with a specialized caching policy aligned to the DGNN execution workflow. Additionally, \sys addresses the challenges of managing structural and temporal dependencies in dynamic graphs through a new distributed training strategy. This approach eliminates communication overheads associated with accessing remote features and redistributing intermediate results. Experimental results demonstrate that \sys achieves up to an order of magnitude speedup compared to state-of-the-art frameworks, tested across various dynamic GNN architectures and real-world graph datasets.

\end{abstract}

%%% Local Variables:
%%% mode: latex
%%% TeX-master: "../main"
%%% End:

\section{Introduction}\label{sec:introduction}
In recent years, there has been a growing interest in Graph Neural Networks 
(GNNs) due to their ability to achieve superior results in a wide range of 
domains, including social networks, knowledge graphs, and medicine~\cite{
PinnerSage2020, Park2019, Lo2018, Stokes2020}. 
To efficiently execute various GNN architectures, state-of-the-art GNN frameworks have implemented numerous optimizations, enabling the processing of large graphs with billions of nodes and edges~\cite{ROC2020, P32021, flexgraph21, GNNLab22, PyG2019, DGLPaper2019, PaGraph2020, Marius2021, GNNAdvisor2021, Dorylus2021}.

While GNNs have achieved remarkable success on static graphs, graphs for real-world applications are inherently \emph{dynamic}, where both the graph structure and node (or edge) features can change over time.
For instance, transportation systems have been collecting massive amounts of data for decades~\cite{metrladataset, pems, trafficbenchmark}, which can be represented as dynamic traffic networks. 
Traffic parameters such as occupancy, volume, and speed are gathered through numerous sensors installed on highways and arterial streets, as associated features of the traffic network. 
A critical task in these systems is predicting future traffic conditions on the road network based on historical data~\cite{li2018dcrnntraffic, STGCN, guo2019attention, huang2020lsgcn, zhang2018gaan}.
Beyond transportation, dynamic graphs play a central role in various domains, such as recommendation systems with constantly changing user preferences~\cite{zhang2022dynamic, song2019session}, anomaly detection in evolving financial systems and social networks~\cite{cai2021structural, pareja2019evolvegcn}, and epidemiological forecasting, as demonstrated during COVID-19~\cite{gnnforcovid, kapoor2020examining}.

\begin{figure}[t]
\centering
\begin{subfigure}[b]{0.4\columnwidth}
   \centering
   \includegraphics[width=0.5\textwidth]{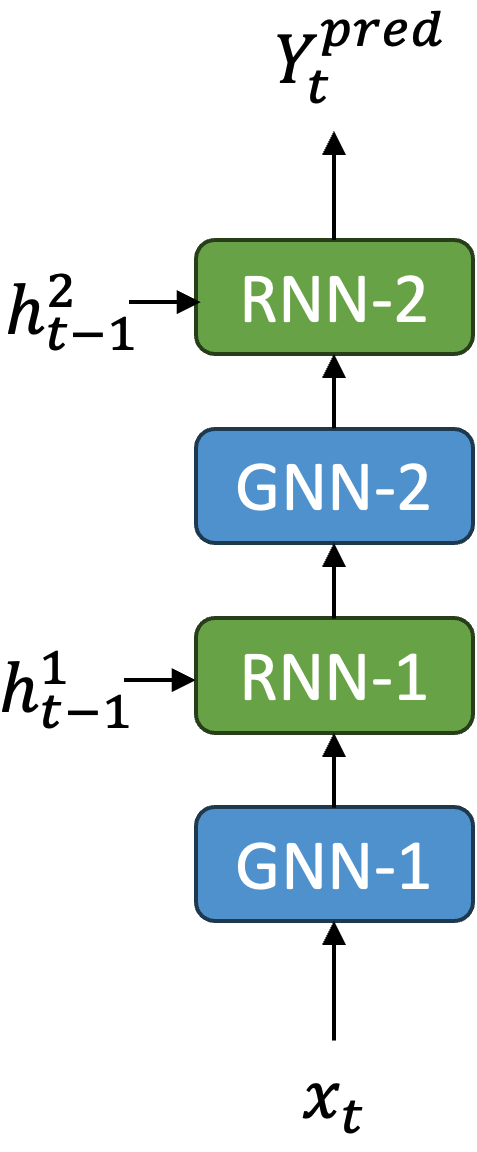}
   \caption{}
   \label{fig:stacked-dgnn} 
\end{subfigure}
\hfill
\begin{subfigure}[b]{0.59\columnwidth}
   \centering
   \includegraphics[width=0.7\textwidth]{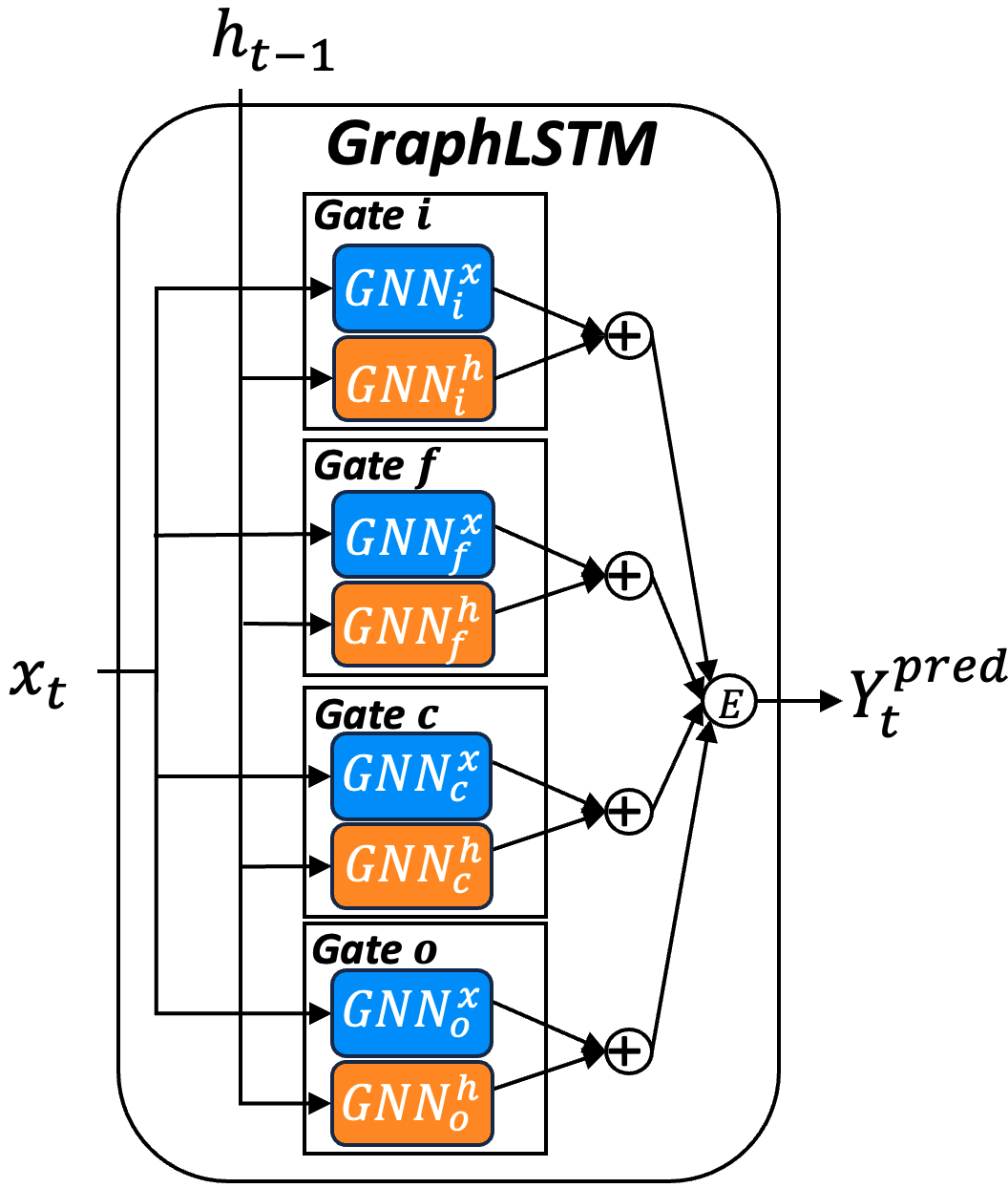}
   \caption{}
   \label{fig:integrated-dgnn}
\end{subfigure}
\caption{(a) A stacked DGNN where an RNN takes
the output of a GNN as its input. (b) An integrated DGNN (GraphRNN) that 
replaces matrix multiplications in RNN gates with GNNs.
}
\label{fig:dgnns}
\end{figure}

To enable these practical applications modeled with dynamic graphs, many dynamic GNNs (DGNNs) have been proposed, demonstrating remarkable success~\cite{
seo2016structured, Manessi2020, li2018dcrnntraffic, chen2021gclstm, 
pareja2019evolvegcn}.
A common approach in DGNNs is to combine \textit{spatial encoding} techniques, which leverage structural information, with \textit{temporal encoding} techniques, which capture the time dependencies in a dynamic graph (\cref{background:dgnn}).
The spatial encoding is enabled by GNNs~\cite{GCN2017, GIN2018, 
GraphSage2017, GAT2018}, whereas temporal encoding typically employs Recurrent 
Neural Networks (RNNs) such as LSTMs~\cite{lstm} or GRUs~\cite{gru}. 
Existing DGNNs combine GNNs and RNNs in two primary ways: by \textit{stacking} them alternately~\cite{seo2016structured, Manessi2020} or by \textit{integrating} their operations into a unified GraphRNN~\cite{seo2016structured, li2018dcrnntraffic, zhang2018gaan, pareja2019evolvegcn, chen2021gclstm, tgcn2020}, as depicted in \cref{fig:dgnns}.

Despite their effectiveness in capturing the dynamic nature of graphs, DGNNs face significant challenges in scaling their training due to \textit{computational complexity} and \textit{communication overhead} (\cref{background:challenges}).
\textit{First}, the combination of GNNs and RNNs introduces substantial redundant computations stemming from graph dynamicity, model architectures, and DGNN execution flow.
To mitigate this, state-of-the-art DGNN systems propose techniques such as parallelizing GNN aggregation operations across multiple snapshots~\cite{DynaGraph, PiPAD} (\cref{background:dgnn-sys}). However, these optimizations often rely on restrictive assumptions about the input graphs, such as small graph sizes that allow parallelization~\cite{PiPAD}, or partial dynamicity, where graph structures~\cite{DynaGraph} or node features~\cite{CBGNN} remain static.

\textit{Second}, achieving efficient DGNN training on large graphs in distributed settings remains an unresolved challenge, with network communication being the primary bottleneck. This bottleneck arises from dependencies inherent in graph structure and time dynamics (\cref{background:dgnn-sys}). Existing solutions attempt to alleviate this by breaking either the structure dependencies~\cite{DynaGraph} or the time dependencies~\cite{ESDGNN} in graph partitioning. However, such compromises limit the generalizability of these approaches, making them suitable only for specific DGNN architectures—stacked or integrated DGNNs—while incurring substantial communication overhead in the other(\cref{background:challenges}).

To this end, we propose \sys, the first system that enables efficient distributed training for both stacked and integrated DGNNs with reuse optimizations and a new distributed training strategy.
\sys is based on three key insights: (1) the unique combination of GNNs and RNNs in DGNNs results in several opportunities to reuse intermediate computations, (2) real-world dynamic graphs change slowly relative to their size, allowing incremental computation to exploit this property, and (3) while DGNNs exhibit time dependencies within each sequence, sequences themselves are independent of one another, enabling opportunities to preserve both the structure dependency and the time dependency in distributed settings.

Based on these insights, \sys is able to significantly accelerate training for both stacked and integrated DGNNs, without any assumption on the underlying dynamic graph.
First, \sys thoroughly explores reuse opportunities inherent in DGNN architectures and their execution flow, minimizing redundant computations (\cref{sec:reuse}).
Second, instead of recomputing each snapshot of the dynamic graph from scratch, \sys employs an incremental approach to compute aggregations, efficiently handling changes across snapshots (\cref{sec:inc-agg}).
To support these reuse optimizations systematically, \sys incorporates a two-level cache store, with a novel cache policy designed to align with DGNN computation patterns, significantly improving cache hit rates and overall performance (\cref{sec:caching}).

% The cache store consists of a global cache where the cached aggregations could be assessed across GraphRNN cells throughout the training process, and a local cache that is designed for accessing in computation of each GraphRNN cell.
% Instead of using traditional caching algorithms, such as least recently used (LRU) and least frequently used (LFU), \sys introduces a new caching policy that is aware of the computation pattern of DGNNs to improve cache hit rate (\cref{sec:caching}).
%and thus achieve better performance (\cref{sec:caching}).

Furthermore, \sys proposes a new approach to distributed DGNN training
that eliminates the overhead of accessing remote features and redistributing intermediate computation results via network communication (\cref{sec:dist-train}).
Specifically, \sys places snapshots as consecutive blocks across machines without breaking the structure and time dependencies, supporting efficient training for general DGNN architectures.
Additionally, the consecutive block placement ensures workload balance and enhances efficiency when combined with reuse and caching techniques in \sys (\cref{sec:partitioning}). 

To further scale training on large graphs, \sys pioneers the application of mini-batch training for DGNNs, ensuring that all data required for the sequence of mini-batch nodes is locally accessible during training, thereby eliminating the overhead of redistributing intermediate results.
Furthermore, \sys redesigns mini-batch training for DGNNs that prioritizes sequence-dimension iteration, significantly improving cache efficiency compared to traditional node-first strategies.
For ease of use, \sys provides a straightforward API with its optimizations under the hood, which can support any new stacked or integrated DGNN architectures(\cref{sec:api}). 

While public large DGNN datasets are scarce, we conduct extensive experiments on large dynamic graphs generated by randomly changing edges (and features) or scaling graph sizes on real graph datasets, following the approach in DGNN systems\cite{DynaGraph} and dynamic graph processing systems\cite{Tegra2021,mo2017gpma,yu2023egraph}.
Our evaluation demonstrates that \sys achieves performance improvements of up to an order of magnitude for both stacked and integrated architectures, compared to best-performing specialized frameworks(\cref{sec:evaluation}).

In summary, we make the following contributions:
\begin{itemize}[noitemsep,topsep=0pt,parsep=0pt,partopsep=0pt]
 \item We identify the limitations of existing solutions in scaling DGNN training for general DGNN architectures.
 \item We introduce \sys, the first system that enables efficient distributed training of stacked and integrated DGNNs.
 \item Experiments show that \sys outperforms the state-of-the-art by up to an order of magnitude on various DGNN architectures and workloads.
\end{itemize}

%%% Local Variables:
%%% mode: latex
%%% TeX-master: "../main"
%%% End:

\section{Background}\label{sec:background}
We begin with an overview of DGNNs(\cref{background:dgnn}), and then discuss existing DGNN training techniques (\cref{background:dgnn-sys}), followed by the motivation and challenges (\cref{background:challenges}).

\subsection{Dynamic Graph Neural Networks}\label{background:dgnn}

\textbf{GNNs vs. DGNNs.} In contrast to static GNNs that compute graph embeddings using only the spatial information, dynamic GNNs (DGNNs) combine both \emph{spatial} and \emph{temporal} information to obtain embeddings of graph entities. This distinction leads to different applications for GNNs and DGNNs: while GNNs are tasked with computing the embedding of the entities in the \emph{present}, DGNNs are typically used for predicting the embedding of entities in the \emph{future}. For instance, DGNNs can be used to forecast how graph entities could evolve. As a result, DGNNs are widely adopted in many critical tasks such as recommendation systems~\cite{RecDGAN2019,zhang2022dynamic} and intelligent transportation systems~\cite{trafficbenchmark, metrladataset, li2018dcrnntraffic}. 
While many techniques for static GNNs, such as \textit{sampling} to address neighborhood explosion \cite{GraphSage2017}, are applicable to the GNN components of a DGNN, the temporal dependencies unique to DGNNs introduce additional challenges.

\textbf{Relation to Dynamic Graph Processing.} 
Dynamic graph processing and DGNNs fundamentally differ in their computational and communication characteristics. Dynamic graph processing algorithms focus on updating algorithmic results in response to graph changes in \textit{the most recent snapshot}. Consequently, in distributed systems, network communication arises solely from the structural dependencies within the current dynamic graph. Conversely, DGNNs are designed to predict future graph snapshots based on sequences of input snapshots, handling both structural dependencies within each snapshot and temporal dependencies across snapshots.
As a result, existing distributed DGNN frameworks require either access to remote features or redistribution of intermediate outputs to manage these dual dependencies (\cref{background:dgnn-sys}).

\textbf{Graph Representation.} Similar to traditional dynamic graph processing systems, there are two representations for a dynamic graph in DGNN frameworks: a discrete representation~\cite{Tegra2021,CBGNN,ESDGNN,DynaGraph,PiPAD} uses a series of snapshots while a continuous representation~\cite{graphbolt19,tesseract2021,TGL,tgcnsys2023} uses timestamps for exact temporal information, such as entity change events and graph streams. 
We focus on discrete dynamic graphs due to their generality and ease of use from both a systems and algorithms perspective. 
For instance, parallelism and batching techniques are easier to apply to snapshots, and extending static GNNs to DGNN on discrete graphs is more straightforward by integrating with a temporal model, e.g., RNN. 
Furthermore, existing studies\cite{Rossetti2018, kazemi2020representation, DGNNSurvey} have shown that discrete dynamic graphs and models are more popular than their continuous counterparts.

\textbf{DGNN Architectures.}
DGNNs combine GNNs to encode structural information, \ie, the neighborhood of a node, with temporal models like RNNs~\cite{lstm, gru} to encode temporal information, \ie, how a node has evolved historically~\cite{DGNNSurvey, rozemberczki2021pytorch}. 
To the best of our knowledge, while a few DGNNs employ attention mechanisms~\cite{sankar2020dysat} for temporal encoding, the combination of GNNs and RNNs remains the most common approach in DGNNs~\cite{trafficbenchmark, DGNNSurvey}.
There are various ways to integrate GNNs and RNNs, but most DGNNs fall into two categories: stacked DGNNs and integrated DGNNs.
\textit{Stacked DGNNs} (\cref{fig:stacked-dgnn}) stack GNNs and RNNs in an alternating fashion~\cite{seo2016structured, Manessi2020}, 
where GNNs handle only structural dependencies, and RNNs are exclusively responsible for temporal dependencies.
\textit{Integrated DGNNs} (\cref{fig:integrated-dgnn}) replace the matrix multiplications in standard RNN cells with GNN operations, creating a GraphRNN cell~\cite{seo2016structured,li2018dcrnntraffic,chen2021gclstm, tgcn2020}. 
In integrated DGNNs, the GNNs within the GraphRNN cell operate on either the input features or hidden states from previous timestamps, making GNN operations both structure- and time-dependent.

\begin{figure}[t]
  \centering
  \includegraphics[width=0.9\linewidth]{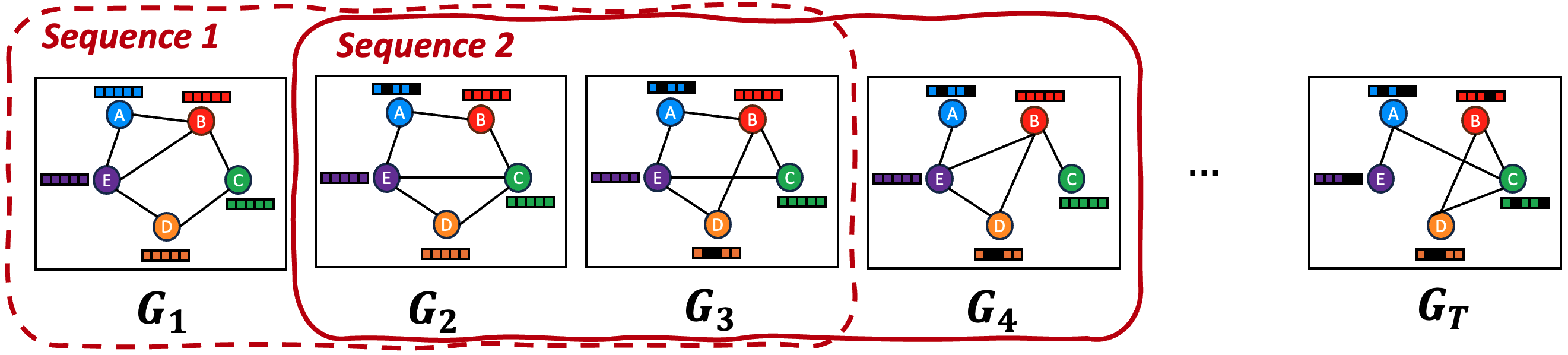}
  \caption{Sliding window mechanism to generate sequences in DGNNs. Consecutive sequences overlap with each other.}
  \label{fig:sliding-window} 
\end{figure}

\textbf{DGNN Execution.} DGNNs ingest a series of historic snapshots as input and output future snapshot(s) as the prediction. 
The input snapshots, referred to as a \emph{sequence}, are typically processed using a sliding window approach (\cref{fig:sliding-window}), where consecutive sequences overlap with one another.
\textit{Each sequence serves as an independent training sample} in DGNNs, where the forward pass computes over the entire sequence of snapshots and the gradients flow backward across time steps of the snapshots during the backward pass.

\begin{figure}[t]
  \centering
  \includegraphics[width=\linewidth]{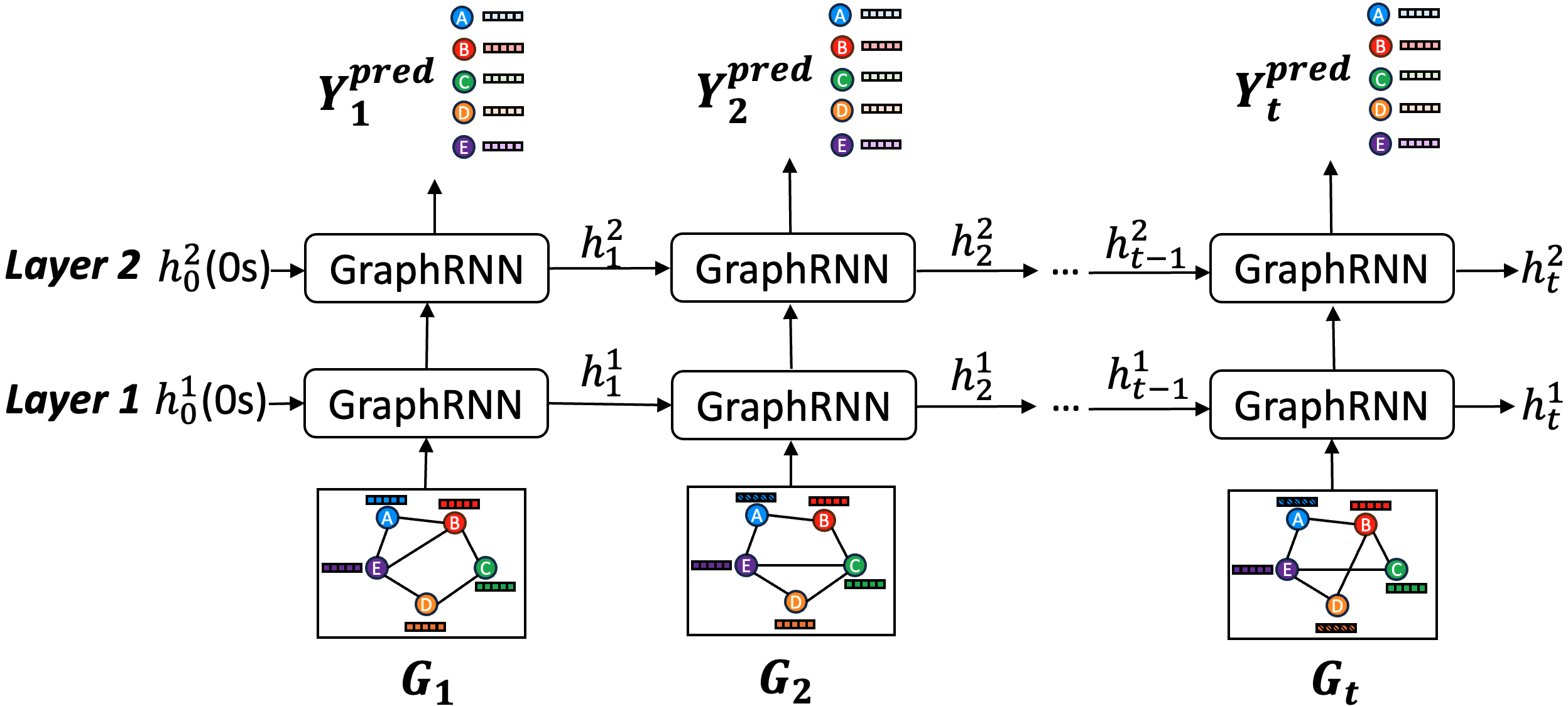}
  \caption{Sequence-to-sequence computation in DGNNs. For each input sequence, the computation of one snapshot $G_t$ depends on the hidden output of the previous snapshot $h_{t-1}$.}
  \label{fig:graph-rnn} 
\end{figure}

Following the execution logic in temporal models, DGNN execution resembles a \textit{sequence-to-sequence} model, as shown in \cref{fig:graph-rnn}. 
To handle variable-length sequences and make predictions based on the entire input sequence, DGNNs leverage the popular \textit{encoder-decoder architecture}~\cite{li2018dcrnntraffic,zhang2018gaan,covid19forecast}, where the encoder and decoder consist of stacked GNN-RNN layers or GraphRNN cells. The encoder encodes the entire input sequence to produce a hidden representation, which serves as input for the decoder to begin predictions.
The decoder operates in a generative and autoregressive manner, using its output predictions as inputs for subsequent timesteps.
Since errors can accumulate over time in this generative architecture, a widely adopted approach is to leverage the ground truth instead of prediction of the last timestep in the decoder to accelerate convergence and enhance performance~\cite{teacherforcing1989, lamb2016professor}, known as the \textit{teacher forcing} strategy.

\begin{figure*}[t]
\centering
\begin{subfigure}{.3\textwidth}
  \centering
  \includegraphics[width=.8\linewidth]{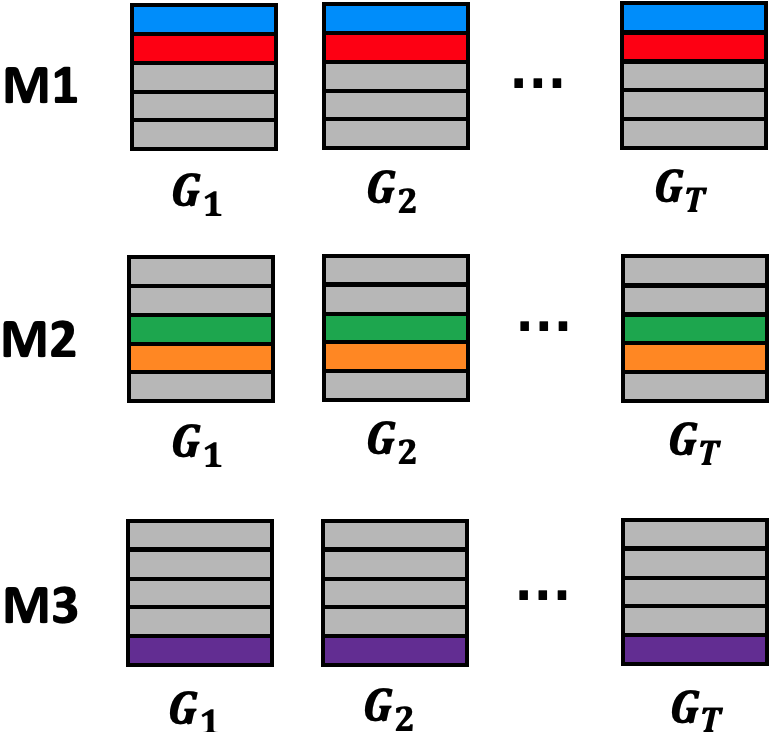}
  \caption{Node partitioning.}
  \label{fig:node-partition}
\end{subfigure}%
\begin{subfigure}{.3\textwidth}
  \centering
  \includegraphics[width=.9\linewidth]{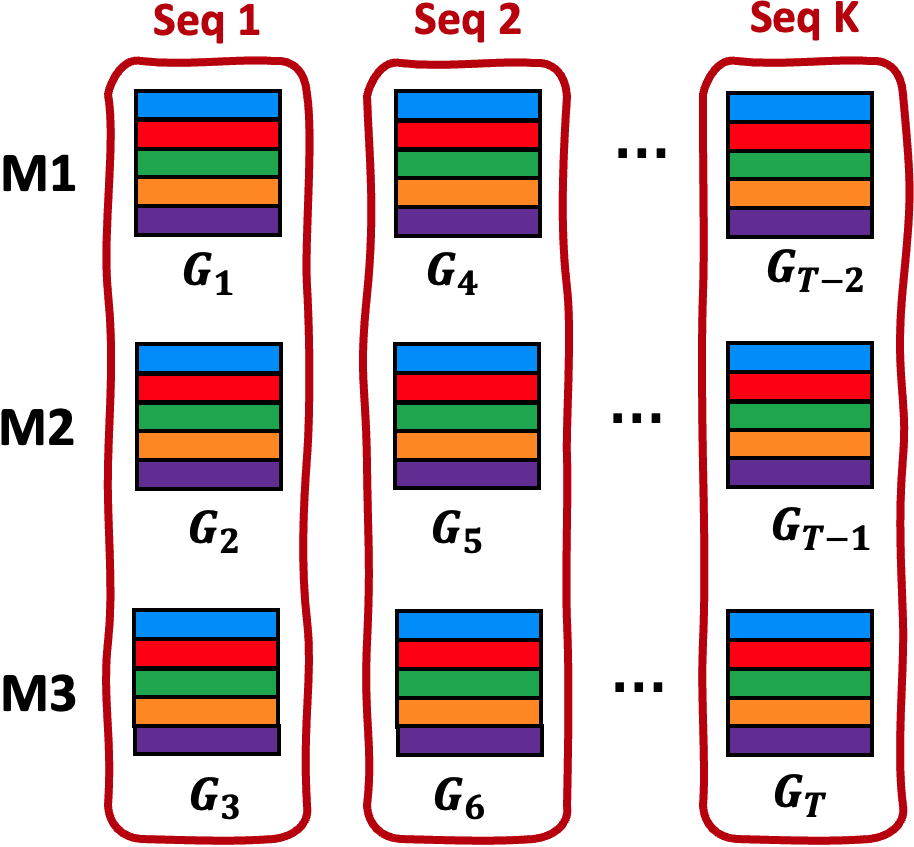}
  \caption{Sequence across machines.}
  \label{fig:esdgnn-partition}
\end{subfigure} %
\begin{subfigure}{.3\textwidth}
  \centering
  \includegraphics[width=.73\linewidth]{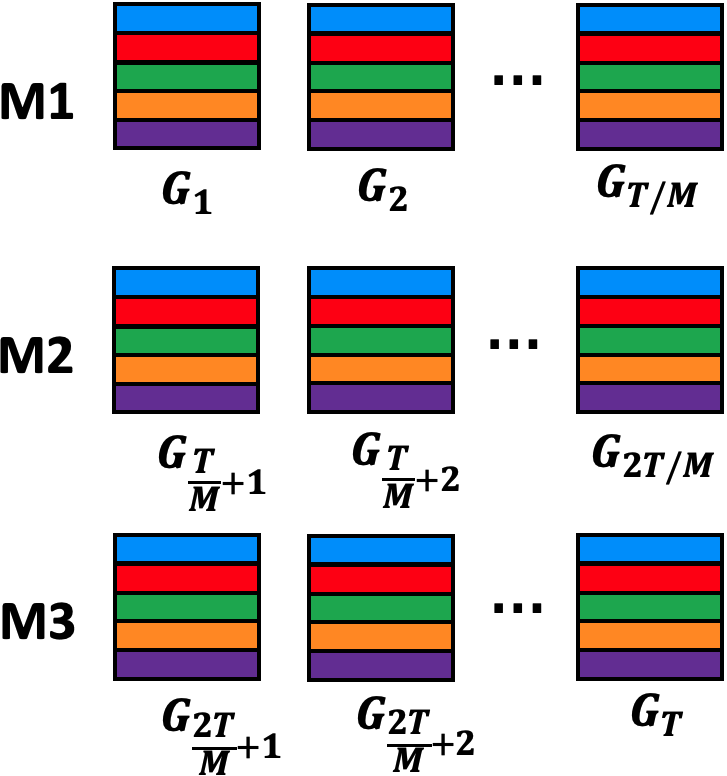}
  \caption{\sys: consecutive blocks across machines.}
  \label{fig:red-partition}
\end{subfigure}
\caption{
  Different graph partitioning schemes for a dynamic graph with $T$ snapshots across $M$ machines. Here, the sequence length is $3$, and each color (or row) represents a node. Features and topology for a node are partitioned on the same machine for all schemes.
}
\label{fig:existing-sys}
\end{figure*}

\subsection{DGNN Frameworks}\label{background:dgnn-sys}

\textbf{Computational Optimizations.}
As discussed in ~\cref{sec:introduction}, many computational optimizations have been explored with assumptions on dynamicity and graph sizes.
CBGNN~\cite{CBGNN} caches intermediate results, assuming that the graph structure evolves with unchanged node features.
DynaGraph~\cite{DynaGraph} focuses on integrated DGNNs and reuses intermediate results across RNN gates. Additionally, for graphs with static graph structure but changing features, DynaGraph proposes timestep fusion to execute graph convolution operation for all snapshots simultaneously, optimized for small graphs.
Similarly, PiPAD~\cite{PiPAD} devised specialized kernels for DGNNs to parallelize the aggregation in the GNN across multiple small snapshots.

\sys's focus on large graphs fundamentally differs from existing work, in which parallelization over multiple snapshots fails to benefit.
Additionally, \sys does not assume partial dynamicity, i.e., both graph structure and features may change over time. 
Besides reuse based on model architecture, e.g., across RNN gates in integrated DGNNs~\cite{DynaGraph}, \sys is the first system to investigate reuse opportunities from the unique DGNN execution flow, 
which benefits both integrated and stacked DGNNs.

\textbf{Distributed DGNN Training.}
To accommodate large graphs in the real world, enabling distributed DGNN training is necessary.
The most intuitive way is to apply node partitioning (\cref{fig:node-partition}) on each snapshot and distribute the GNN computations across multiple machines, as in static GNNs.
For GNN computation, each machine pulls remote features of the node neighborhood from other machines for each snapshot.
Once the GNN results are available for all snapshots(timesteps) in the sequence, each machine can execute the RNN independently for a subset of nodes.
However, intelligent node partitioning schemes, e.g., METIS~\cite{METIS} used in DGL~\cite{DGLPaper2019}, incur either high computation overhead or high memory overhead even for a static graph~\cite{P32021}, not to mention that a dynamic graph has many snapshots.
Besides, existing partitioning schemes~\cite{karypis1998fast, gonzalez2012powergraph, zhang2016exploring} for static graphs are not directly applicable to dynamic graphs. 
If applying the partitioning scheme on the first snapshot and subsequent snapshots follow the same partitioning, it quickly loses its benefits as the graph evolves. 
Hence, existing DGNN systems instead use random node partitioning for simplicity~\cite{DynaGraph}.

To alleviate communication overhead, ESDGNN~\cite{ESDGNN} places the snapshots in a sequence across machines (\cref{fig:esdgnn-partition}) so that the GNN computation can proceed in parallel. 
However, the RNN computation has to operate on an entire sequence, and thus the GNN output has to be \textit{redistributed} via network communication such that each machine can compute an entire sequence for a subset of nodes. 
Even then, it is much superior to partitioning each snapshot, because the communication volume of node partitioning increases with the number of machines, while the communication volume of redistributing GNN output is fixed with the hidden size.
However, this approach cannot efficiently support \textit{integrated DGNNs} (\cref{fig:integrated-dgnn}), where GNNs operate on both input features and hidden vectors from the previous timestep, which makes GNN operations have time dependency and thus cannot be parallelized.

In summary, existing DGNN systems optimize for a certain group of models and target small graphs.
However, supporting various DGNN models for large graphs is necessary to enable real-world applications, which is the focus of \sys.

\subsection{Motivation and Challenges}\label{background:challenges}
\begin{table}[h]
\centering
\small
\addtolength{\tabcolsep}{-0.2em}
\begin{tabular}{ccccc}
 \toprule
      & \multicolumn{2}{c}{\textbf{Stacked DGNN}} & \multicolumn{2}{c}{\textbf{Integrated DGNN}} \\
  \hline
      & \textbf{Comm.(GB)} & \textbf{Time(s)} & \textbf{Comm.(GB)} & \textbf{Time(s)} \\
 \midrule
  \textbf{DGL}       & 1070 & 1824 & 1070 & 6138 \\
  \textbf{DynaGraph} & 1333 & 2510 & 1333 & 3688 \\
  \textbf{ESDGNN}    &  230 & 1185  &  359 & 22474 \\
  \textbf{\sys}      &   0  &  389  &   0  & 1273 \\
\hline
\end{tabular}
\caption{Comparison of epoch communication volume and time across DGL\cite{DGLPaper2019}, DynaGraph\cite{DynaGraph}, ESDGNN\cite{ESDGNN}, and \sys. 
Gradient synchronization communication volumes, which are identical across all frameworks, are excluded for clarity. 
}
\label{table:motivation}
\end{table}

While existing DGNN systems incorporate techniques to reduce computation and communication overhead (\cref{background:dgnn-sys}), scaling the training of diverse DGNN architectures on large graphs remains unresolved. As shown in \cref{table:motivation}, node partitioning (\cref{fig:node-partition}) used by DGL and DynaGraph incurs substantial communication overhead for remote feature access in both stacked and integrated DGNNs. 
While DynaGraph offsets part of its communication overhead through computational optimizations for integrated DGNNs, the optimizations cannot be generalized to stacked DGNNs.
Conversely, ESDGNN’s sequence partitioning approach (\cref{fig:esdgnn-partition}) mitigates communication for stacked DGNNs but supports integrated DGNNs poorly due to time-dependent GNN operations across machines, which hinders parallelization.

To address these limitations, we propose a novel distributed training scheme that places consecutive blocks across machines (\cref{fig:red-partition}). 
Our key observation is that while DGNNs exhibit time dependencies within each sequence, sequences are independent training samples of DGNNs.
Such a graph placement supports sliding-window execution naturally and preserves structural dependencies within snapshots and temporal dependencies within each sequence, thereby eliminating remote feature access and intermediate state redistribution.

\textbf{Challenges.} While placing consecutive blocks across
machines minimizes communication overhead, we identify the following challenges to efficiently scale DGNN training.

\textit{Redundant computation}. Computing every snapshot from scratch as in existing DGNN frameworks incurs significant computational redundancy as real-world graphs change slowly relative to their size. This problem is exacerbated in the case of a sliding window, where the overlap results in many computations to repeat. 
Moreover, the combination of GNNs and RNNs often involves repeated aggregation over the same data.
Hence, there is a critical need for systematic exploration of all these reuse opportunities to enhance efficiency.

\textit{Limited GPU memory.} 
While the host machine memory can hold its partition in memory, it is infeasible for devices (GPUs) to compute full snapshots of large graphs.
Hence, supporting mini-batch training that accommodates graph dynamicity and the DGNN execution flow is essential.

\textit{General support.} Many optimizations and training techniques in existing frameworks are under strong assumptions of small graphs or static features (\cref{background:dgnn-sys}). 
A DGNN framework must generalize to support both stacked and integrated architectures on graphs with dynamic structures and features.

%%% Local Variables:
%%% mode: latex
%%% TeX-master: "../main"
%%% End:

\section{DGNN Training with \sys}\label{sec:system}

To enable efficient and scalable training for DGNNs, we propose \sys to solve the above challenges with reusing optimizations (\cref{sec:reuse} \& \cref{sec:inc-agg}), DGNN-aware caching (\cref{sec:caching}), and a novel distributed training scheme (\cref{sec:dist-train}).

\subsection{Three Reuse Opportunities} \label{sec:reuse}
Existing frameworks perform aggregation for the same snapshot repeatedly due to 
the structural GNN-RNN combination, sequence execution as well as the encoder-decoder architecture.
However, the overlapped snapshots across sequences also bring three reuse 
opportunities.

\textit{First}, each gate of a GraphRNN contains two GNN operations in an integrated 
DGNN (\cref{fig:integrated-dgnn}), where one is for features and the other 
for hidden states. 
For all GNN operations in a GraphRNN that operate on features, aggregation 
results are the same for a snapshot and similarly for GNNs on hidden 
states. Hence, we can aggregate only once for input features and hidden state, respectively, and 
reuse them across gates for each GraphRNN. 
\textit{Second}, the sequence execution in DGNNs due to the sliding window mechanism 
(\cref{fig:sliding-window}) results in many overlapped snapshots across 
consecutive sequences.
This brings another opportunity to reuse since the aggregation for the overlapped snapshots has already been performed at the previous sequence execution. %(\filledcircledtikz{2})
\textit{Last}, with the teacher-forcing strategy, the decoder takes ground truths as inputs (\cref{background:dgnn}).
The ground truths are often future snapshots, 
which will present in the subsequent input sequences.
Therefore, we can reuse the aggregation results from the decoder when the encoder takes 
those snapshots as inputs. 
To systematically support these reuse optimizations, we build a two-level cache that efficiently stores and evicts aggregations (\cref{sec:caching}).

Reuse operations across gates is a distinct feature of integrated DGNNs, whereas reuse across sequences and between the encoder and decoder is broadly applicable to general DGNNs, as both stacked and integrated architectures incorporate sequence execution and the encoder-decoder framework. 
Moreover, reuse across gates can be applied to both inputs and hidden states at each layer, while the other two reuse techniques can only be applied to inputs, and thus only the first layer.
Fortunately, the first GNN layer is the most computationally intensive, given that input graph features are typically high-dimensional, ranging from hundreds to several thousands of dimensions~\cite{OGB2020, PinSage2018, AGL, P32021}. In contrast, hidden dimensions in later layers are significantly smaller, usually limited to several tens~\cite{P32021, DGLPaper2019, DynaGraph, ESDGNN}.
Hence, reusing aggregations across sequences and model parts in the first-layer GNN can bring substantial speedups as shown in \cref{fig:inc-reuse}. 

\subsection{Incremental Aggregation}\label{sec:inc-agg}
\begin{figure}[t]
    \centering
    \includegraphics[width=\linewidth]{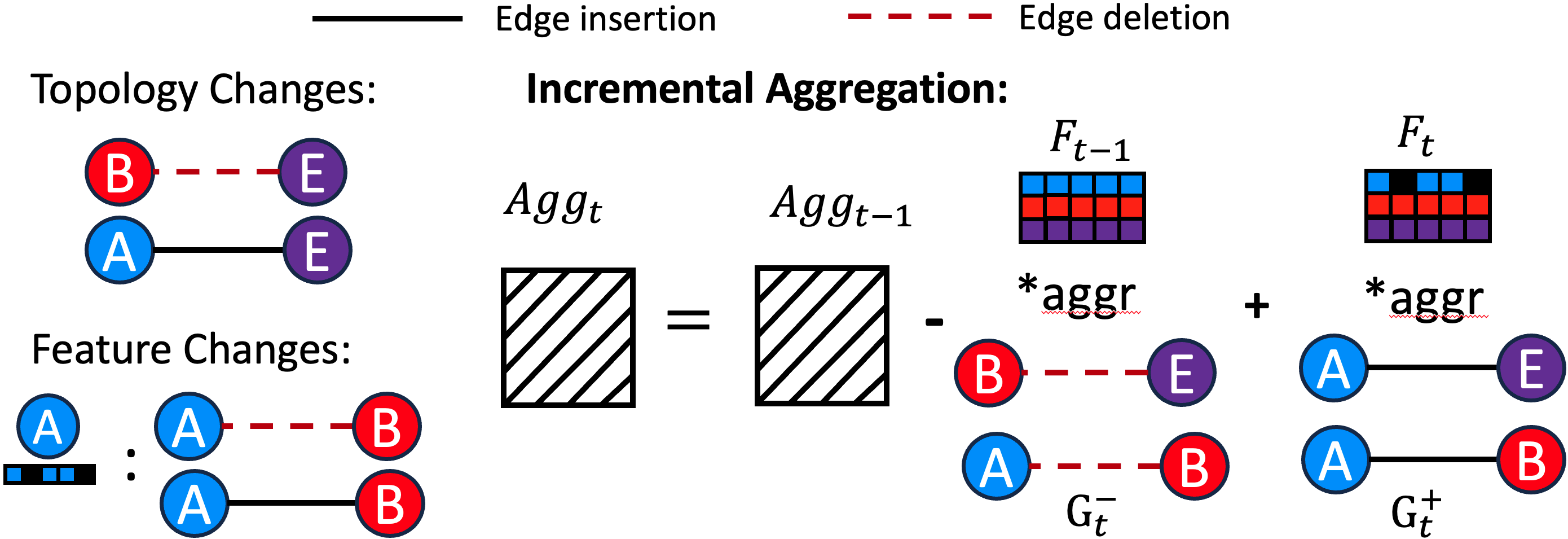}
    \caption{\sys extracts two delta graphs $G^-_t$ (edge deletions) and $G^+_t$ (edge insertions) and computes aggregation results incrementally to eliminate redundant computation.}
    \label{fig:inc-agg}
\end{figure}

Existing DGNN frameworks perform graph aggregation step for each snapshot \emph{from scratch} in a dynamic graph, 
which is inefficient since snapshots change slowly relative to 
their size.
Moreover, some DGNNs\cite{pareja2019evolvegcn, TM-GCN} tend to smoothen the differences across the snapshots to improve the stability of the prediction, which further reduces the change ratio across snapshots.
Hence, in \sys, we use \emph{incremental aggregation} for snapshots, which minimizes redundant computation resulting from the large common parts across 
snapshots. 

To illustrate incremental computations, we use an example that shows 
how we calculate aggregation for snapshot $G_{t}$ in \cref{fig:inc-agg}. 
From $t$ to $t+1$, in terms of topology change, there is an edge deletion 
($\circledtikz{B}\rightarrow\circledtikz{E}$) and an edge insertion 
($\circledtikz{A}\rightarrow\circledtikz{E}$). Moreover, features of node \circledtikz{A} are also changed.

To enable incremental aggregation, the key is to figure out what contribute to 
aggregation result at the current timestep, compared with the aggregation results 
from the previous timestep.
It is relatively straightforward to handle topology changes. Intuitively, if 
there is an edge deletion from a source node $src$ to a target node $tgt$, 
previous features of $src$ need to be removed from $tgt$'s aggregation; likewise, if there is an edge insertion from $src$ to $tgt$, the current features of 
$src$ should be added to $tgt$'s aggregation. 
Moreover, feature changes can also be incorporated as edge deletions and 
insertions. 
Since a node's features only contribute to aggregations of its out-neighbors, 
a node's feature change can be viewed as edge deletions from it to its 
out-neighbors at previous timestep and edge insertions from it to its 
out-neighbors at current timestep.
Putting them together, for each snapshot $G_t$, we reflect both structure changes and feature changes as two delta graphs $G^-_t$ and $G^+_t$, where $G^-_t$ consists of edge deletions and $G^+_t$ consists of edge insertions.

During training, \sys can utilize $G^-$ and $G^+$ to perform aggregation 
incrementally.
As shown in \cref{fig:inc-agg}, instead of aggregating snapshot at $t$ 
from scratch:
\begin{equation}
     Agg_{t} = F_{t} *_{aggr} G_{t}
\end{equation}
where $*_{aggr}$ means aggregation operation while $Agg_{t}$, $F_{t}$ and 
$G_{t}$ represent aggregation results, features, and topology for snapshot 
$t$, respectively.
\sys incrementally performs aggregation at $t$ using the previous 
results $Agg_{t-1}$ by:
\begin{equation}
     Agg_{t} = Agg_{t-1} - F_{t-1} *_{aggr} G^-_{t} + F_{t} *_{aggr} G^+_{t}
\end{equation}
Note that both $G^-$ and $G^+$ are very small relative to the size of the snapshot 
in large-scale dynamic graphs.
Hence, \sys can eliminate redundant computation resulting from the large 
common parts across snapshots.
In case the change ratio is higher than a user-defined threshold for certain snapshots where the size of $G^-$ 
and $G^+$ are comparable to the size of the snapshot, \sys falls back 
to the aggregation-from-scratch approach for those. 

\sys’s choice to focus on edge changes is based on the constraints of existing DGNNs that assume constant matrix sizes within a sequence because of the need to capture temporal changes of fixed-size representations in temporal models like RNNs. However, it is trivial to extend our design to node additions and deletions by introducing nodes that do not have edges based on the maximum number of nodes in the graph. 

We note that for aggregation functions like $sum()$,
we can directly apply the above procedure for incremental computation.
However, for aggregation functions such as $max()$ and $min()$,
edge deletions cannot be incrementally computed as we do not store 
all its old contributions.
For such functions, \sys additionally stores $edge ids$ of edges that contribute to the old aggregations.
Upon edge deletions of those old contributions, \sys falls back to the aggregation-from-scratch approach.
Otherwise, incremental aggregation can be performed.
Optionally, users can also choose to cache all the contribution values instead of only the final aggregation results, which saves the re-aggregation overhead but needs more memory for caching.
For weighted aggregation that involves learnable weights of the neural network, e.g., graph attention neural networks\cite{GAT2018}, \sys performs incremental aggregation between two backward passes and aggregates from scratch after each update of weights.
Built-in aggregation functions supported by \sys include $sum()$, $mean()$, $max()$, and $min()$, which capture the majority of GNN architectures. %used in DGNNs.

\subsection{Two-level Cache Store}\label{sec:caching}
\sys utilizes a two-level cache store to enable low-overhead DGNN training with its reuse optimizations.
The cache store is composed of a global cache and a local cache, where cached aggregations in the global cache can be assessed across layers throughout the training process while the local cache is designed for each layer.
Instead of using traditional eviction algorithms, such as LFU and LRU, \sys's caching policy is aware of the execution flow of DGNN, such as the sliding-window mechanism and teacher-forcing strategy, to improve cache hit rate and thus achieve better performance.

\subsubsection{Global Cache}
The global cache is built for storing intermediate aggregation results of input features, which can be reused across gates, sequences, and parts of the model (\cref{sec:reuse}).
With teacher forcing enabled, the input features of a snapshot $G_t$ are aggregated for the first time at the first GraphRNN layer of the decoder and \sys puts $Agg^{input}_t$ in the global cache. 
Since each GraphRNN has several gates, for example, $3$ and $4$ gates for GRU and LSTM, respectively, $Agg^{input}_t$ is computed only once for the first gate and reused from the global cache for the rest gates.
Later, $Agg^{input}_t$ can be reused again from the global cache when snapshot $G_t$ is in the input sequence for the encoder, where reuse across gates is also applicable.
Due to the sliding window mechanism that generates overlapped sequences, snapshot $G_t$ can appear in multiple consecutive sequences, which further increases the reuse of $Agg^{input}_t$ in the global cache.
Moreover, the global cache also facilitates incremental aggregation as discussed in \cref{sec:inc-agg}.
During the incremental computation of $Agg_t$, \sys gets cached aggregation $Agg_{t-1}$ from the global cache, which is put in the global cache the first time aggregating the input feature of snapshot $G_{t-1}$.

\subsubsection{Local Cache}
For each GraphRNN layer, a local cache is employed to store aggregation results for intermediate hidden vectors, including those from the last timestep and from the last layer.
Let $h^l_t$ be the hidden vector for layer $l$ at timestep $t$.
The GraphRNN cell at layer $l$ takes $h^{l-1}_t$ and $h^{l}_{t-1}$ as inputs (\cref{fig:graph-rnn}) and the aggregations for them over $G_t$ are needed for each gate in the GraphRNN.
\sys aggregates $h^{l-1}_t$ and $h^{l}_{t-1}$ over snapshot $G_{t}$ only once, which are cached and reused across all gates. 

\begin{figure*}[t]
  \centering
  \includegraphics[width=0.9\linewidth]{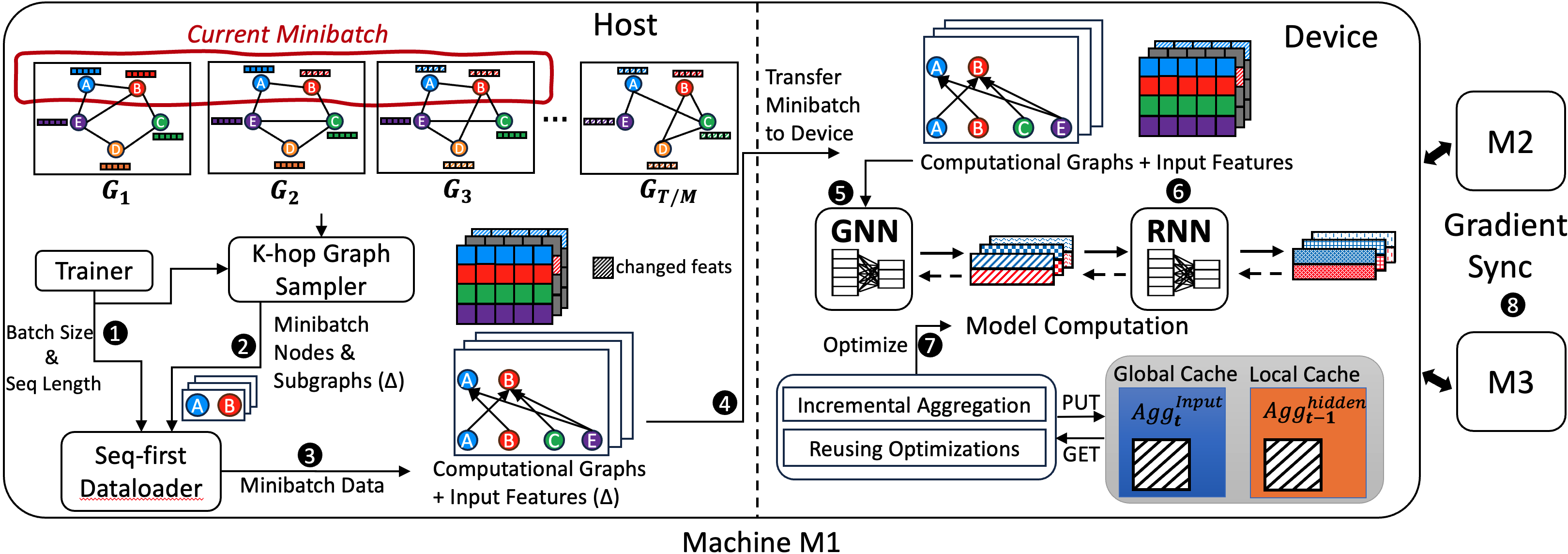}
  \caption{Overview of distributed DGNN execution in \sys (details in \cref{sec:dgnn-execution}). 
  }
  \label{fig:red-dist} 
\end{figure*}

\subsubsection{Eviction Policy}
We adopt a scoring strategy to assign numerical scores to each aggregation with higher values representing aggregation results are more beneficial to the cache.
One key observation in DGNN training is that, unlike other workloads where the access pattern of objects is unpredictable because of a lack of dependency among objects, \textit{DGNN access to aggregations of snapshots is fully predictable}.
Unlike other caching algorithms, such as LFU and LRU, which focus on access history, \sys focuses on predictable future access to score aggregations.
Inspired by GDSF\cite{GDSF}, which uses $Freq(f)/Size(f)+Clock$ to score a file $f$, where $Freq$ is the frequency count, $Size$ is the file size and $Clock$ is a global aging factor which gives more recent files higher scores.
Specifically, \sys considers three metrics to score aggregations: future access count, imminence (how soon it will be accessed in the future), and size.
Following GDSF, we assign equal weights to each of these factors and score each aggregation:
\begin{equation}
    Priority(Agg) = \frac{F(Agg)}{S(Agg)} - I(Agg),
\end{equation}
where $F$ is the future access count, $S$ is the size and $I$ is the imminence.
We now discuss how to calculate future access count and imminence. 
Without loss of generality, we illustrate how to calculate the future access count with the common scenario where an LSTM-based DGNN predicts a sequence $q+1$ based on the last sequence $q$.
Let $D$ be the number of GraphRNN layers, $K$ be the number of gates in each GraphRNN, $g$ be the current gate number, $S$ be the stride of the sliding window, $L$ be the sequence length, and $idx$ be the index of the snapshot in the current sequence.

\textit{\textbf{Future access count.}}
When aggregation $Agg$ is computed and cached during \textit{a GraphRNN layer other than the first layer}, only local cache is used within the GraphRNN computation.
We have $F(Agg) = K - g$, where the future access count equals the rest number of gates in the GraphRNN.
When the aggregation $Agg$ is cached during a GraphRNN computation \textit{at the first layer in encoder}, we have $F(Agg) = max(1, 2*idx-2*S+1)*K -g$, considering future access count in the encoder as sliding window moves and future access count in the decoder as teacher state.
When the aggregation $Agg$ is cached during a GraphRNN computation \textit{at the first layer in decoder}, similarly we have $F(Agg) = max(2, 2*idx-2*S+2)*K -g$.

\textit{\textbf{Imminence.}}
To account for the time that aggregation is accessed from now on, we count the number of timesteps that will be passed (computed) without using the aggregation as imminence.
The sooner a cached aggregation is accessed, the lower the imminence it has. 
Since each GraphRNN reuses aggregation results across gates, the local aggregation will be accessed and reused immediately after caching, which gives $I(Agg) = 0$.
For global cache, if $Agg$ is computed and cached in the encoder, it first needs to finish the rest $L-idx$ snapshots in the sequence and the next access time is in the decoder after $idx-S$ snapshots, and together we have $I(Agg) = L - S$.
Similarly, if $Agg$ is computed in the decoder, the next access time is in the encoder and we have $I(Agg) = L - 1$.

\textit{\textbf{Expiration.}}
\sys also employs an additional eviction policy based on the expiration of aggregation.
When the future access count of an aggregation goes down to $0$, we mark it as expired and evict it directly without considering its caching score, as it will not be accessed in the future.
This prompt eviction policy enables \sys to further reduce the memory overhead for caching the aggregations.

\subsection{Distributed Training}\label{sec:dist-train}
To minimize computational redundancy, \sys employs a new graph placement strategy and re-designs mini-batch training to support large graphs for DGNNs, as depicted in~\cref{fig:red-dist}.
\sys utilizes model checkpointing to recover from failures, acknowledging that failures are common in distributed systems. However, this work does not specifically focus on failure recovery mechanisms.

\subsubsection{Graph Placement}\label{sec:partitioning}
The key insight enabling \sys to minimize communication overhead is that \textit{time dependencies among snapshots exist only within each sequence and not across sequences}.
As shown in \cref{fig:red-partition}, \sys splits the training dynamic graph into consecutive blocks based on the number of machines $M$.
Each machine is assigned $T/M$ consecutive snapshots, where $T$ is the total number of snapshots in the discrete dynamic graph.
Similarly, \sys places input features associated with each snapshot on the same machine correspondingly.
Consecutive block placement allows \sys to achieve load balance in DGNN training by ensuring that each machine processes an equal number of sequences and batches, where the sequence length and batch size are consistent throughout training.

For sequences that are placed across machines, \sys supports two solutions.
If a machine has enough storage and memory to hold additional snapshots, \sys supports overlapped snapshot placement, where sequences are placed on each machine in their entirety overlapping a few snapshots with other machines.
Otherwise, \sys retrieves remote snapshots from other machines during training as needed.

We note that, compared to existing partitioning schemes illustrated in \cref{fig:node-partition} and \cref{fig:esdgnn-partition}, \sys’s graph placement results in partitions that need comparable disk storage and host memory, for storing and loading the partitions, as all schemes partition the graph evenly across machines.
Similar to existing distributed GNN and DGNN systems, e.g., DGL~\cite{DGLPaper2019}, P3~\cite{P32021} and ESDGNN~\cite{ESDGNN}, \sys is an in-memory DGNN system, where the host(CPU) maintains the data partition in memory, while each device(GPU) has limited memory that is insufficient to process a sequence of full snapshots for large graphs. 
We defer supporting out-of-core graphs to future work.

\subsubsection{DGNN Execution}\label{sec:dgnn-execution}
After graph placement, each machine owns a local consecutive block of snapshots.
To support large dynamic graphs, \sys enables mini-batch training for DGNNs, where each mini-batch contains a set of nodes $N^b$ and a sample is a sequence of these nodes, i.e. $N^b_t, N^b_{t+1}, ..., N^b_{t+L-1}$, where $L$ is the sequence length.
Given a batch size and the sequence length(\filledcircledtikz{1}), each machine generates a sequence of computational graphs for the nodes in the minibatch(\filledcircledtikz{2}), as shown in \cref{fig:red-dist}.
For the first snapshot, computational graph generation follows the standard process used in static GNNs~\cite{DGLPaper2019,P32021,understandGNN22}, by pulling the k-hop neighborhood of each node.
When neighborhood sampling is enabled for scalability and load balancing, \sys retrieves the sampled k-hop neighborhood; otherwise, it retrieves the entire k-hop neighborhood.
For subsequent snapshots, \sys optimizes memory usage and host-device data transfers by extracting only the relevant graph changes between snapshots.
After that, the input features needed for the minibatch are extracted accordingly (\filledcircledtikz{3}).

To start execution, \sys first transfers the samples from the hosts to the devices (\filledcircledtikz{4}) and each machine can independently execute its sample in a data-parallel fashion.
In the forward pass, for GNN executions, \sys executes the computational graphs for each snapshot without incurring any network communication (\filledcircledtikz{5}), as each device possesses the complete graph structure and input features required for the mini-batch.
After GNN execution, the RNN takes the output of the GNN as input.
Since each device already holds a complete sequence of mini-batch nodes, \sys can immediately proceed with RNN computation (\filledcircledtikz{6}), instead of redistributing GNN outputs to construct a sequence for RNNs as in existing systems(\cref{background:dgnn-sys}).
As shown in \cref{fig:red-dist}, the device on machine $M1$ processes GNN outputs of the sequence for mini-batch nodes \circledtikz{A} (blue) and \circledtikz{B} (red), executing RNN operations locally without any network communication.
During the training, the two-level cache store serves incremental aggregation and reuse optimizations to minimize computational redundancy(\filledcircledtikz{7}). 
As \sys incurs no data dependencies across machines in the forward pass, each machine can perform the backward pass independently and invokes only global gradient synchronization at each layer boundary (\filledcircledtikz{8}). 

\subsubsection{Rethinking Mini-batch Training}\label{sec:mini-batch}
For static GNNs, mini-batch training is typically performed by iterating over nodes in the graph.
Intuitively, we can follow this tradition to enable mini-batch training for DGNNs.
Given a sequence of snapshots, we select a set of nodes and extract the computational graphs, and together this sequence of computational graphs for the selected nodes is a sample.
To generate the next sample, we select another set of nodes and repeat the above process.
After iterating over all the nodes in the graph for the current sequence, the sliding window moves to the next sequence, and the same procedure is repeated.

However, we observe that this mini-batch training procedure results in significant memory overhead and a low cache hit rate when \sys’s optimizations are enabled. This is because \sys needs to cache intermediate results for all nodes in the current sequence, allowing these results to be reused until computations for the next sequence begin.

To address this challenge, \sys introduces a novel mini-batch training logic for DGNNs. 
Instead of iterating over nodes and then moving to the next sequence (\textit{node-first}) as in static GNNs, \sys proposes iterating over all sequences for a mini-batch of nodes before sampling the next mini-batch (\textit{seq-first}).
Specifically, for a given mini-batch of nodes, \sys processes all the sequences assigned to the current machine by advancing the sliding window through the sequences until the last one. 
Once all sequences for the mini-batch are processed, the sliding window resets, and the next mini-batch of nodes is sampled.
This new approach reduces memory overhead by caching aggregations only for the current mini-batch, while also improving performance by increasing the cache hit rate within \sys’s cache store. 

\subsection{\sys API}\label{sec:api}
\begin{python}{Using \sys’s API to implement a GCRN-M2 model with encoder-decoder structure.}{lst:api}
class GCN(nn.module):
  def __init__(in_feats, out_feats):
    linear = nn.Linear(in_feats, out_feats)
    relu = nn.ReLU()
  # generate messages
  def msg_udf(g, feats, u, v): return msg
  # reduce aggregated messages from scratch
  def reduce_udf(agg_msg): return sum(agg_msg)
  # compute new representation
  def update(msg_rst):
    return relu(linear(msg_rst))
  def forward(t, delta_g, feats):
    msg_rst = msg_pass(t, delta_g, feats, msg_udf, reduce_udf)
    return update(msg_rst)

gcrn = integrate(GraphLSTM, GCN)
seq2seq_gcrn = stack_seq_model(in_feats, out_feats, num_layers, gcrn, teacher_forcing=True)
\end{python}

\sys provides a simple API that allows developers to easily integrate its optimizations into new DGNN architectures. 
Listing ~\ref{lst:api} outlines how GCRN-M2~\cite{seo2016structured} can be 
implemented in \sys. 
By leveraging \sys’s API, developers can construct DGNNs with ease, while the system automatically handles optimizations discussed in ~\cref{sec:system}.
To define a GCN layer, the developer starts by implementing the \emph{msg_udf} function, which generates messages based on the incoming source node representations, and the \emph{reduce_udf} function, which aggregates the neighborhood representations using summation.
In the \emph{forward} function, the \emph{msg_pass} method takes the delta graph as input, performing incremental aggregation internally as described in \cref{sec:inc-agg}. 
The intermediate outputs from the message-passing step are then passed into a fully connected layer using the \emph{update} function. 
After defining the GCN model, the developer can subsequently invoke the \emph{integrate} function to combine the GCN with LSTM, thereby creating a DGNN layer.
If an encoder-decoder structure is needed, the \emph{stack_seq_model} function can be utilized to stack the DGNN layers into an encoder-decoder architecture. 

%%% Local Variables:
%%% mode: latex
%%% TeX-master: "../main"
%%% End:

\section{Implementation}\label{sec:implementation}

We developed a prototype implementation of \sys’s techniques, as described in \cref{sec:system}, leveraging the Deep Graph Library (DGL)~\cite{DGLPaper2019} for neighborhood sampling and aggregation, and PyTorch as the neural network execution backend.
We implement on top of DGL as it is one of the most widely used static GNN systems, while techniques in \sys are extensible to any static GNN system, as they are agnostic to the specific implementation of GNN execution.
First, we implemented a two-level cache store with the proposed cache policies to support incremental processing and reuse optimizations. This required modifying DGL’s message-passing primitives to handle incremental aggregation for dynamic graphs. These modified primitives compute the aggregation for the current timestep using delta aggregations internally, enhancing both memory and computational efficiency.
Second, we replaced DGL’s \emph{DataLoader}, which is designed for static graph training by iterating over a set of nodes or edges, with a \emph{SeqDataLoader} that iterates over snapshots and creates sequence samples for DGNN training. Under \emph{delta} mode, the \emph{SeqDataLoader} constructs and transfers only the minimal data needed for incremental DGNN computation, reducing overhead and improving performance.
Third, to support efficient distributed DGNN training, we implemented \sys’s snapshot placement strategy, preserving both structure and time dependencies. 
Finally, we wrapped user-defined DGNN models using PyTorch’s \emph{DistributedDataParallel()} module to synchronize and update model weights.
\section{Evaluation}\label{sec:evaluation}
We evaluate \sys on four representative DGNNs across four large dynamic graphs, compared with state-of-the-art systems including DynaGraph~\cite{DynaGraph} and ESDGNN~\cite{ESDGNN}. Our evaluation shows that:
\begin{itemize}
  \item \sys achieves superior performance compared to state-of-the-art DGNN systems, outperforming DynaGraph and ESDGNN by 2.9-12.8$\times$ and 2.8-17.7$\times$, respectively. 
  \item The reuse and incremental optimizations proposed by \sys result in up to 5.4$\times$ speedup, reducing memory usage by up to 73\% with the two-level cache store.
  \item \sys scales gracefully with the number of machines, features and hiddens, sequence length, and change ratio.
\end{itemize}

\textbf{Experimental Setup:} Our distributed experiments were conducted 
on a GPU cluster with 8 nodes, each with dual AMD Epyc 7513 CPUs, 512 GB of RAM, and one Nvidia Tensor Core A100 40GB GPU. 
GPUs on the same node are connected via a shared PCIe interconnect, and nodes 
are connected via a 10 Gbps Ethernet interface. 
Servers run 
with CUDA library v11.8, PyTorch v1.13.0, DGL v0.9.

\begin{table}[t!]
\centering
\small
\begin{tabular}{cccc}
 \toprule
 \textbf{Graph} & \textbf{Nodes} & \textbf{Edges} & \textbf{Features}\\
 \midrule
  PEMS-BAY-LARGE~\cite{li2018dcrnntraffic} & 1.3 M & 11M & 128 \\
  METR-LA-LARGE~\cite{metrladataset} & 1.6 M & 14.1M & 128 \\
  OGB-Products~\cite{ogbdataset} & 2.4 M & 123.7 M & 100 \\
  OGB-Papers~\cite{ogbdataset} & 111 M & 1.6 B & 128 \\
\hline
\end{tabular}
\caption{Graph datasets used in evaluating \sys. $Nodes$ ($Edges$) column indicate the nodes (edges) in \emph{each} snapshot.}
\label{table:datasets}
\end{table}

\textbf{Datasets:} 
We list the four datasets used in our experiments in \cref{table:datasets}, where the statistics are provided for \textit{each snapshot}.
The first two datasets, METR-LA-LARGE~\cite{metrladataset} and PEMS-BAY-LARGE~\cite{li2018dcrnntraffic}, represent dynamic traffic graphs collected from sensors on highways in Los Angeles County and the Bay Area, respectively. Real-world transportation systems have deployed millions of traffic sensors~\cite{metrladataset, pems}, with the quantity continuously increasing over time. However, the publicly released graphs only include a few hundred nodes (sensors)\cite{li2018dcrnntraffic}. 
Hence, we scale these datasets by replicating their original versions multiple times, following the approach used in existing distributed DGNN systems\cite{DynaGraph}.
The latter two datasets, OGB-Product and OGB-Papers, are part of the Open Graph Benchmark (OGB)~\cite{ogbdataset}. OGB-Product is an Amazon product co-purchasing network, while OGB-Papers is a citation graph, which are the two largest static graphs in the benchmark. To convert these real-world static graphs into their dynamic counterparts, we generate 100 snapshots by randomly modifying edges (with equal probabilities for deletion and insertion) and updating node features (assigning new features of the same dimensions as the initial ones), using a uniform distribution for change ratios between $0\%$ and $100\%$.
Additionally, we evaluate the scaling characteristics of \sys with increasing fixed change ratios in \cref{sec:eval-scaling}.

\textbf{Baselines:} \label{para:baselines}
DynaGraph~\cite{DynaGraph} incorporates a simple distributed training approach using random node partitioning with reuse-across-gates optimization, thus allowing our comparison by enabling the corresponding optimization in \sys.
We also developed a prototype implementation of ESDGNN~\cite{ESDGNN}, a distributed DGNN system optimized for stacked DGNNs. Specifically, we implemented its distributed training scheme by placing snapshots within each sequence across machines (\cref{background:dgnn-sys}).
We note that existing DGNN systems including DynaGraph~\cite{DynaGraph} and ESDGNN~\cite{ESDGNN} only support full graph training, which cannot train the evaluated large graphs.
Therefore, we port the mini-batch training techniques in \sys to their prototypes.

\textbf{Models:} We use four different DGNN models:
GCRN-M1~\cite{seo2016structured}, CD-GCN~\cite{Manessi2020}, GCRN-M2~\cite{seo2016structured}, and TGCN~\cite{tgcn2020}. 
These models represent the state-of-the-art dynamic GNNs in various tasks:
GCRN-M1 and CD-GCN are stacked DGNNs that stack a GNN and an RNN in an alternative way, while GCRN-M2 and TGCN are integrated DGNNs. 
They differ in which GNN and/or which RNN (LSTM~\cite{lstm} or GRU~\cite{gru}) they use.
For stacked DGNNs, we use 2 stacked GNN-RNN pairs in GCRN-M1 and CD-GCN; for integrated DGNNs, we use the encoder-decoder structure, where the encoder and decoder have 2-layer GraphRNNs, respectively. 
Unless mentioned otherwise, we enable a [25, 10] neighborhood
sampling for all DGNNs, following the default setting in the state-of-the-art~\cite{GraphSage2017,P32021}, where a maximum of 25 and 10 neighbors are sampled for the first and second hops, respectively.
We use a hidden size of 64 for all models and set the sequence length to 8, which are middle-of-the-pack settings. 
Mini-batch size is set to 10,000 in all our experiments.
For experiments that evaluate the impact of varying configurations 
(e.g., features), we use a middle-of-the-pack dataset in terms of size 
(OGB-Products).
Without losing generality, we evaluate our cache store using the model GCRN-M2 since an integrated model covers all resuing opportunities so that we can demonstrate the effectiveness of \sys's techniques.

\subsection{Overall Performance}\label{sec:eval-overall}

\begin{figure}
  \centering
  \input{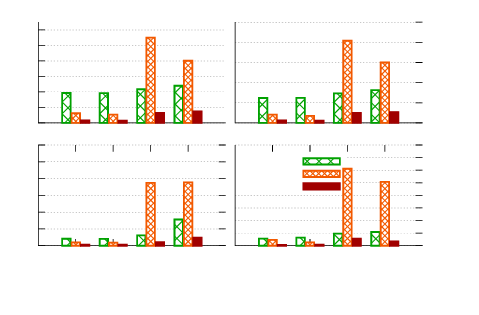}
  \caption{
    Overall, \sys's speedups range up to 12.8$\times$ and 17.7$\times$ compared to DynaGraph and ESDGNN, respectively.
  }
  \label{fig:overall-dist-new}
\end{figure}

We first compare \sys's overall performance to DynaGraph and ESDGNN, two state-of-the-art DGNN systems.
As shown in \cref{fig:overall-dist-new}, \sys can significantly outperform the baselines for both stacked and integrated DGNNs over all the evaluated graphs by 2.9-12.8$\times$ and 2.8-17.7$\times$ compared to DynaGraph and ESDGNN, respectively. 

For stacked DGNNs, 
ESDGNN performs better than DynaGraph by avoiding remote access to input features, but it still incurs communication overhead by redistributing intermediate results and gradients across the machines. 
Instead, \sys incurs no communication by splitting and placing snapshots as consecutive blocks.
Together with computational optimizations of reuse techniques, \sys achieves the best performance by eliminating redundant overhead.

For integrated DGNNs, \ie, GCRN-M2 and TGCN, 
ESDGNN performs worst among the baselines because it places snapshots within each sequence across machines and each machine is blocked by other machines' computation due to time dependency.
We notice that the speedup of \sys for integrated DGNNs is higher than that of stacked DGNNs.
This is because more reuse opportunities, \ie, reuse across gates and reuse in teacher-forcing, exist in integrated DGNNs, compared with stacked ones.
Moreover, the LSTM-based models (GCRN-M2) benefit more from \sys's techniques compared to GRU-based models (TGCN), due to the increased number of gates in the LSTM-based models.

\subsection{Impact of Sampling}\label{sec:eval-sampling}

\begin{figure}
  \centering
  \input{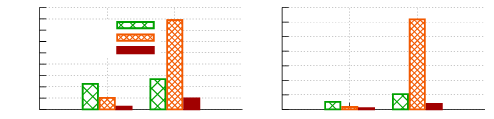}
  \caption{Without sampling, \sys maintains its benefits while baselines struggle to handle larger graphs.}
  \label{fig:inc-reuse}
\end{figure}

To evaluate \sys’s performance when the underlying task does not support sampling, we disable sampling and repeat the experiments detailed in \cref{sec:eval-overall}. Without sampling, most baselines face significant challenges in training the largest dataset (OGB-Papers). Specifically, DynaGraph experiences high memory overhead when pulling computational graphs, and ESDGNN requires an extremely long training time for integrated DGNNs.
For METR-LA-LARGE, the epoch times remain similar with or without sampling. However, for OGB-Products, the epoch times increase significantly without sampling. This difference arises because METR-LA-LARGE is relatively sparse, with low node degrees (8 on average), while OGB-Products is dense, with higher node degrees (25 on average), making it more sensitive to the absence of sampling. 
\sys demonstrates the ability to maintain its benefits without sampling across both architectures and graphs. In contrast, the baselines face difficulties in handling large graphs, highlighting the advantage of \sys’s approach in preserving both structural and temporal dependencies during DGNN training.

\subsection{Impact of Reusable Computation}\label{sec:eval-reuse-inc}
\begin{figure}
  \centering
  \input{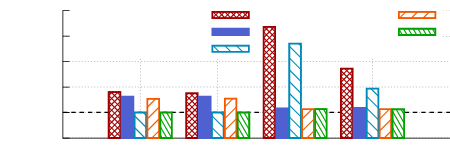}
  \caption{
    Impact of \sys’s reusable optimizations. 
    }
  \label{fig:inc-reuse}
\end{figure}

To evaluate the impact of each reusable optimization, we report the speedup achieved when enabling individual design choices in \cref{fig:inc-reuse}.
Overall, integrated DGNNs (e.g., GCRN-M2 and T-GCN) demonstrate greater speedup compared to stacked DGNNs (e.g., GCRN-M1 and CD-GCN). This is because stacked DGNNs do not leverage \emph{reuse across gates} or \emph{reuse in teaching} with their simpler structure and execution flow.
Among integrated DGNNs, \emph{reuse across gates} delivers the highest speedup, as its computational savings scale proportionally with the number of RNN units and gates.
We note that the reported individual speedups are not additive or multiplicative, because a single cached aggregation can support multiple optimizations simultaneously, such as \emph{reuse across gates} and \emph{reuse across sequence}, contributing to overlapping performance benefits.

\subsection{Cache Store Evaluation}
\begin{figure}
  \centering
  \input{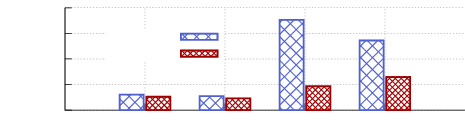}
  \caption{
    \sys's cache store reduces memory usage significantly by avoiding generating redundant aggregations.
  }
  \label{fig:cache-memory}
\end{figure}

\noindent \textbf{Memory Usage}
As shown in \cref{fig:cache-memory}, our cache store does not introduce additional memory overhead but significantly reduces GPU memory usage during training. 
This is because of \sys’s ability to reuse aggregations, eliminating redundant intermediate results that would otherwise require additional memory. 
Moreover, \sys minimizes caching memory usage with the automatic eviction of expired aggregations, aligned with the DGNN execution flow.
The memory savings for integrated DGNNs are substantially greater than those for stacked DGNNs due to the higher number of reuse opportunities, including reuse across gates and model parts(\cref{sec:reuse}).

\begin{figure}
  \centering
  \input{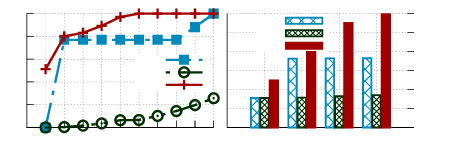}
  \caption{
    \sys's cache policy significantly outperforms LRU and LFU for DGNN training.
  }
  \label{fig:cache-policy}
\end{figure}

\noindent \textbf{Effectiveness of Caching Policy} 
We evaluate the effectiveness of \sys’s DGNN-aware cache policy under limited cache memory and compare it with two traditional algorithms, LRU and LFU. 
As shown in \cref{fig:cache-policy}, \sys achieves a higher cache hit rate and greater training throughput, demonstrating its advantage in understanding the DGNN execution flow.
Among the baselines, LRU outperforms LFU in DGNN training, primarily because the seq-first mini-batch training aligns well with LRU. Aggregations for the current snapshot (most recently used) can be accessed and reused immediately across RNN gates, benefiting LRU. In contrast, LFU retains older aggregations with higher past access frequency, even when they can no longer be reused.

However, LRU remains suboptimal compared to \sys. With a small cache-to-data size (e.g., $10\%$), \sys achieves a $52\%$ cache hit rate, while LRU delivers $0\%$. This is because such a small cache size can only store aggregations for either input or hidden states. During gate computations in GraphRNN cells, input and hidden state aggregations are accessed alternately, causing LRU to consistently fail to provide the required aggregations. In contrast, \sys’s cache policy, being aware of DGNN access patterns, effectively caches the correct aggregations.
As the cache-to-data size increases from $10\%$ to $20\%$, LRU begins to support reuse across gates by storing at least two sets of aggregations for inputs and hidden states. However, between $20\%$ and $80\%$, LRU’s cache hit rate stagnates at $78\%$ because it cannot support reuse across sequences and model parts, where aggregations needed for multiple timestamps away are evicted.
For cache-to-data sizes exceeding $80\%$, LRU gradually reaches $100\%$ cache hit rate as the cache size becomes sufficient to store all relevant aggregations. In contrast, \sys achieves a $100\%$ hit rate with $60\%$ cache-to-data size by intelligently retaining necessary aggregations and promptly evicting redundant ones, demonstrating its superior caching strategy tailored to DGNN workflows.

\begin{figure}[t]
  \centering
  \input{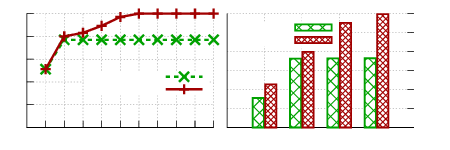}
  \caption{
    \sys's \textit{seq-first} strategy increases cache hit rate and throughput, compared with the standard \textit{node-first} strategy.
  }
  \label{fig:cache-mini-batch}
\end{figure}

\noindent \textbf{Impact of \sys's Mini-batch Training}
As shown in \cref{fig:cache-mini-batch}, \sys’s seq-first strategy for DGNN training significantly improves both cache hit rate and throughput, compared to the standard node-first approach.
With a small cache-to-data size ($10\%$ to $20\%$), both strategies achieve similar cache hit rates. However, as cache size increases, the cache hit rate of the node-first strategy stagnates because it progresses to new mini-batch nodes, preventing the reuse of cached aggregations across sequences. Conversely, the seq-first strategy enables immediate access and reuse of cached aggregations for the current mini-batch nodes in subsequent sequences, leading to a progressively higher cache hit rate and throughput.

\subsection{Scaling Characteristics}\label{sec:eval-scaling}
\begin{figure}
  \centering
  \input{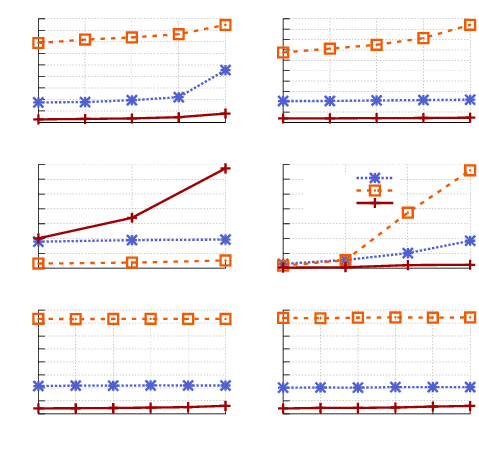}
  \caption{
    \sys’s strong scaling with increased feature (hidden) dimension, machines, sequence length, and change ratio.
  }
  \label{fig:scaling}
  % \Description{
  %   Figure of \sys’s scaling characteristics.
  % }
\end{figure}

We conduct comprehensive experiments to evaluate scaling properties of \sys in \cref{fig:scaling}.
First, we vary the size of the feature dimension and hidden dimension.
It shows that \sys exhibits near-linear scaling characteristics and maintains its benefits with the increase of \emph{feature and hidden dimension size}. 
While DynaGraph maintains its performance varying hidden size, its performance degrades as the size of feature dimension increases, due to the large communication overhead to exchange features.
In contrast, ESDGNN scales better than DynaGraph with the increase of feature dimension size but fails to maintain its benefit as hidden dimension size increases, due to the large network communication volume needed for redistributing intermediate results and gradient.
Since \sys avoids both redundant communication due to graph structure and time dependencies, the overhead brought by increased feature or hidden dimensions is small.

Next, we investigate the scaling characteristics with the increase in the \emph{number of machines}. 
We find that \sys exhibits near linear scaling characteristics, while DynaGraph and ESDGNN’s throughput scales poorly.
This is because GPU resources in DynaGraph and ESDGNN get held back by overwhelming data communication, while \sys can effectively reduce this overhead with its graph placement scheme without breaking structure and time dependencies.

We further analyze how \sys performs with the increased \emph{sequence length}, which decides how many snapshots are fed into the model as a single sample.
As shown in \cref{fig:scaling}, as the sequence length increases, the average epoch time of DynaGraph and ESDGNN increases drastically due to computation overhead proportional to the sequence length.
However, \sys maintains its benefits because of its reuse techniques that avoid unnecessary computations as sequence length increases.

We also present the scaling characteristics varying in \emph{change ratio of features and edges}.
As features and edges change more noticeably across snapshots, epoch time on \sys increases while DynaGraph and ESDGNN maintain their performance. This is because the baselines compute aggregations from scratch and the change ratio does not affect computation time. In contrast, \sys computes aggregations incrementally where the computation time increases with a larger change part. However, since real-world graphs change slowly relative to their size, \sys can benefit from its incremental aggregation for most cases. Note that the change ratio \textit{only affects one of \sys’s optimizations}, i.e., incremental aggregation. 

\subsection{Correctness}\label{sec:eval-correctness}
\begin{figure}
  \centering
  \input{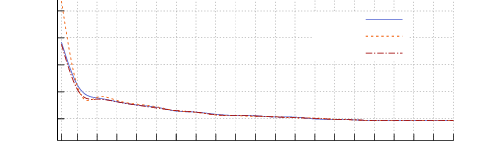}
  \caption{
    Correctness of \sys’s optimizations and distributed training strategy.
  }
  \label{fig:accuracy}
\end{figure}

Finally, we evaluate the correctness of our approach in \sys on METR-LA-LARGE.
We present results on this dataset due to its real-world dynamicity, minimizing the influence of random changes and providing a more accurate reflection of \sys’s correctness. 
We also evaluate the convergence curves on all other datasets, which align consistently with those of the baselines.
We report the test Mean Absolute Error (MAE) for the prediction task in 100 epochs for \sys, DynaGraph, and ESDGNN.
As shown in \cref{fig:accuracy}, \sys can achieve the same MAE as DynaGraph and ESDGNN, thus ensuring its correctness while being able to gain $2.9\times$ and $8.1\times$ speedup, compared to DynaGraph and ESDGNN, respectively.

%%% Local Variables:
%%% mode: latex
%%% TeX-master: "../main"
%%% End:

\section{(Other) Related Work}\label{sec:related}

\paraf{(Dynamic) Graph Processing Systems:}
Enabling graph algorithms on a large scale, several systems can process
massive graphs even with trillions of edges~\cite{Powerlyra2015, khandelwal2017zipg, TrillionEdges, Tegra2021, gpl2021}. In
graph processing and graph mining literature, there has been an increasing interest in processing
dynamic, or \emph{time-evolving} graphs, for the purpose of temporal, historic, 
or real-time analysis~\cite{CuSha2014,ASAP2018,graphbolt19,graphone19,gong2021automating, tesseract2021}. These systems focus on enabling graph algorithms
and graph mining and do not support embedding generation as required in GNNs. 

\paraf{Static GNN Systems:}
GNNs have become an active area of research in the recent past, both in the
machine learning and the systems community. 
The ML community has
proposed many GNN architectures, each with its advantages for particular tasks~\cite{Hamilton2017, SpectralGCN2016, GCN2017, GAT2018}.
Recently, many efforts have been made to scale
GNNs to large-scale graphs by leveraging accelerators such as GPUs~\cite{GRIP2020}. 
Some
of these proposals are single-machine systems, while others support distributed settings. PyTorch Geometric~\cite{PyG2019} and DGL~\cite{DGLPaper2019} are popular frameworks optimized for GNN training. 
GNNLab~\cite{GNNLab22} adopts a factored design for a multi-GPU setup, where each GPU is dedicated to either graph sampling or model training.
FlexGraph~\cite{flexgraph21}, ROC~\cite{ROC2020}, and $P^3$ ~\cite{P32021} propose techniques to scale distributed GNN training with different graph partitioning and parallelism schemes.
However, none of them support dynamic GNN training in an efficient way. 

%%% Local Variables:
%%% mode: latex
%%% TeX-master: "../main"
%%% End:

\section{Conclusion}\label{sec:conclusion}
In this paper, we investigated the problem of efficient and scalable DGNN training.
We observed that the redundant computation and high communication overhead hinder DGNN systems from scaling well and analyzed their root causes.
Hence, we presented \sys, a system that supports scalable DGNN training, and showed that \sys significantly outperforms the state of the art across extensive experiments. 

%%% Local Variables:
%%% mode: latex
%%% TeX-master: "../main"
%%% End:

\phantomsection
\label{EndOfPaper}

\bibliographystyle{plain}
\bibliography{main}

\begin{thebibliography}{10}

\bibitem{pems}
{Performance Measurement System (PeMS) Data Source}.
\newblock \url{https://dot.ca.gov/programs/traffic-operations/mpr/pems-source}.

\bibitem{tesseract2021}
Laurent Bindschaedler, Jasmina Malicevic, Baptiste Lepers, Ashvin Goel, and Willy Zwaenepoel.
\newblock Tesseract: distributed, general graph pattern mining on evolving graphs.
\newblock In {\em Proceedings of the Sixteenth European Conference on Computer Systems}, EuroSys '21, page 458–473, New York, NY, USA, 2021. Association for Computing Machinery.

\bibitem{cai2021structural}
Lei Cai, Zhengzhang Chen, Chen Luo, Jiaping Gui, Jingchao Ni, Ding Li, and Haifeng Chen.
\newblock Structural temporal graph neural networks for anomaly detection in dynamic graphs.
\newblock In {\em Proceedings of the 30th ACM International Conference on Information \& Knowledge Management}, CIKM '21, page 3747–3756, New York, NY, USA, 2021. Association for Computing Machinery.

\bibitem{ESDGNN}
Venkatesan~T. Chakaravarthy, Shivmaran~S. Pandian, Saurabh Raje, Yogish Sabharwal, Toyotaro Suzumura, and Shashanka Ubaru.
\newblock Efficient scaling of dynamic graph neural networks.
\newblock In {\em Proceedings of the International Conference for High Performance Computing, Networking, Storage and Analysis}, SC '21, New York, NY, USA, 2021. Association for Computing Machinery.

\bibitem{chen2021gclstm}
Jinyin Chen, Xueke Wang, and Xuanheng Xu.
\newblock Gc-lstm: Graph convolution embedded lstm for dynamic link prediction, 2021.

\bibitem{Powerlyra2015}
Rong Chen, Jiaxin Shi, Yanzhe Chen, and Haibo Chen.
\newblock Powerlyra: Differentiated graph computation and partitioning on skewed graphs.
\newblock In {\em Proceedings of the Tenth European Conference on Computer Systems}, EuroSys '15, New York, NY, USA, 2015. Association for Computing Machinery.

\bibitem{GDSF}
Ludmila Cherkasova.
\newblock {\em Improving WWW proxies performance with greedy-dual-size-frequency caching policy}.
\newblock Hewlett-Packard Laboratories Palo Alto, CA, USA, 1998.

\bibitem{TrillionEdges}
Avery Ching, Sergey Edunov, Maja Kabiljo, Dionysios Logothetis, and Sambavi Muthukrishnan.
\newblock One trillion edges: Graph processing at facebook-scale.
\newblock {\em Proc. VLDB Endow.}, 8(12):1804--1815, August 2015.

\bibitem{gru}
Kyunghyun Cho, Bart van Merrienboer, {\c{C}}aglar G{\"{u}}l{\c{c}}ehre, Fethi Bougares, Holger Schwenk, and Yoshua Bengio.
\newblock Learning phrase representations using {RNN} encoder-decoder for statistical machine translation.
\newblock {\em CoRR}, abs/1406.1078, 2014.

\bibitem{SpectralGCN2016}
Micha\"{e}l Defferrard, Xavier Bresson, and Pierre Vandergheynst.
\newblock Convolutional neural networks on graphs with fast localized spectral filtering.
\newblock In D.~Lee, M.~Sugiyama, U.~Luxburg, I.~Guyon, and R.~Garnett, editors, {\em Advances in Neural Information Processing Systems}, volume~29, pages 3844--3852. Curran Associates, Inc., 2016.

\bibitem{PyG2019}
Matthias Fey and Jan~E. Lenssen.
\newblock Fast graph representation learning with {PyTorch Geometric}.
\newblock In {\em ICLR Workshop on Representation Learning on Graphs and Manifolds}, 2019.

\bibitem{P32021}
Swapnil Gandhi and Anand~Padmanabha Iyer.
\newblock P3: Distributed deep graph learning at scale.
\newblock In {\em 15th {USENIX} Symposium on Operating Systems Design and Implementation ({OSDI} 21)}, pages 551--568. {USENIX} Association, July 2021.

\bibitem{gong2021automating}
Shufeng Gong, Chao Tian, Qiang Yin, Wenyuan Yu, Yanfeng Zhang, Liang Geng, Song Yu, Ge~Yu, and Jingren Zhou.
\newblock Automating incremental graph processing with flexible memoization.
\newblock {\em Proceedings of the VLDB Endowment}, 14(9):1613--1625, 2021.

\bibitem{gonzalez2012powergraph}
Joseph~E Gonzalez, Yucheng Low, Haijie Gu, Danny Bickson, and Carlos Guestrin.
\newblock Powergraph: Distributed graph-parallel computation on natural graphs.
\newblock In {\em Presented as part of the 10th $\{$USENIX$\}$ Symposium on Operating Systems Design and Implementation ($\{$OSDI$\}$ 12)}, pages 17--30, 2012.

\bibitem{DynaGraph}
Mingyu Guan, Anand~Padmanabha Iyer, and Taesoo Kim.
\newblock Dynagraph: Dynamic graph neural networks at scale.
\newblock In {\em Proceedings of the 5th ACM SIGMOD Joint International Workshop on Graph Data Management Experiences \& Systems (GRADES) and Network Data Analytics (NDA)}, GRADES-NDA '22, New York, NY, USA, 2022. Association for Computing Machinery.

\bibitem{guo2019attention}
Shengnan Guo, Youfang Lin, Ning Feng, Chao Song, and Huaiyu Wan.
\newblock Attention based spatial-temporal graph convolutional networks for traffic flow forecasting.
\newblock In {\em Proceedings of the AAAI conference on artificial intelligence}, volume~33, pages 922--929, 2019.

\bibitem{GraphSage2017}
Will Hamilton, Zhitao Ying, and Jure Leskovec.
\newblock Inductive representation learning on large graphs.
\newblock In I.~Guyon, U.~V. Luxburg, S.~Bengio, H.~Wallach, R.~Fergus, S.~Vishwanathan, and R.~Garnett, editors, {\em Advances in Neural Information Processing Systems}, volume~30, pages 1024--1034. Curran Associates, Inc., 2017.

\bibitem{Hamilton2017}
William~L. {Hamilton}, Rex {Ying}, and Jure {Leskovec}.
\newblock {Representation Learning on Graphs: Methods and Applications}.
\newblock {\em IEEE Data Engineering Bulletin}, page arXiv:1709.05584, September 2017.

\bibitem{lstm}
Sepp Hochreiter and Jürgen Schmidhuber.
\newblock {Long Short-Term Memory}.
\newblock {\em Neural Computation}, 9(8):1735--1780, 11 1997.

\bibitem{OGB2020}
Weihua {Hu}, Matthias {Fey}, Marinka {Zitnik}, Yuxiao {Dong}, Hongyu {Ren}, Bowen {Liu}, Michele {Catasta}, and Jure {Leskovec}.
\newblock {Open Graph Benchmark: Datasets for Machine Learning on Graphs}.
\newblock {\em arXiv e-prints}, page arXiv:2005.00687, May 2020.

\bibitem{ogbdataset}
Weihua Hu, Matthias Fey, Marinka Zitnik, Yuxiao Dong, Hongyu Ren, Bowen Liu, Michele Catasta, and Jure Leskovec.
\newblock Open graph benchmark: Datasets for machine learning on graphs.
\newblock {\em CoRR}, abs/2005.00687, 2020.

\bibitem{tgcnsys2023}
Chengying Huan, Shuaiwen~Leon Song, Yongchao Liu, Heng Zhang, Hang Liu, Charles He, Kang Chen, Jinlei Jiang, and Yongwei Wu.
\newblock T-gcn: A sampling based streaming graph neural network system with hybrid architecture.
\newblock In {\em Proceedings of the International Conference on Parallel Architectures and Compilation Techniques}, PACT '22, page 69–82, New York, NY, USA, 2023. Association for Computing Machinery.

\bibitem{huang2020lsgcn}
Rongzhou Huang, Chuyin Huang, Yubao Liu, Genan Dai, and Weiyang Kong.
\newblock Lsgcn: Long short-term traffic prediction with graph convolutional networks.
\newblock In {\em IJCAI}, volume~7, pages 2355--2361, 2020.

\bibitem{ASAP2018}
Anand~Padmanabha Iyer, Zaoxing Liu, Xin Jin, Shivaram Venkataraman, Vladimir Braverman, and Ion Stoica.
\newblock {ASAP}: Fast, approximate graph pattern mining at scale.
\newblock In {\em 13th {USENIX} Symposium on Operating Systems Design and Implementation ({OSDI} 18)}, pages 745--761, Carlsbad, CA, October 2018. {USENIX} Association.

\bibitem{Tegra2021}
Anand~Padmanabha Iyer, Qifan Pu, Kishan Patel, Joseph~E. Gonzalez, and Ion Stoica.
\newblock {TEGRA}: Efficient {Ad-Hoc} analytics on evolving graphs.
\newblock In {\em 18th USENIX Symposium on Networked Systems Design and Implementation (NSDI 21)}, pages 337--355. USENIX Association, April 2021.

\bibitem{metrladataset}
H.~V. Jagadish, Johannes Gehrke, Alexandros Labrinidis, Yannis Papakonstantinou, Jignesh~M. Patel, Raghu Ramakrishnan, and Cyrus Shahabi.
\newblock Big data and its technical challenges.
\newblock {\em Commun. ACM}, 57(7):86–94, jul 2014.

\bibitem{ROC2020}
Zhihao Jia, Sina Lin, Mingyu Gao, Matei Zaharia, and Alex Aiken.
\newblock Improving the accuracy, scalability, and performance of graph neural networks with roc.
\newblock In I.~Dhillon, D.~Papailiopoulos, and V.~Sze, editors, {\em Proceedings of Machine Learning and Systems}, volume~2, pages 187--198, 2020.

\bibitem{kapoor2020examining}
Amol Kapoor, Xue Ben, Luyang Liu, Bryan Perozzi, Matt Barnes, Martin Blais, and Shawn O'Banion.
\newblock Examining covid-19 forecasting using spatio-temporal graph neural networks.
\newblock {\em arXiv preprint arXiv:2007.03113}, 2020.

\bibitem{covid19forecast}
Amol Kapoor, Xue Ben, Luyang Liu, Bryan Perozzi, Matt Barnes, Martin Blais, and Shawn O'Banion.
\newblock Examining {COVID-19} forecasting using spatio-temporal graph neural networks.
\newblock {\em CoRR}, abs/2007.03113, 2020.

\bibitem{METIS}
George Karypis and Vipin Kumar.
\newblock A fast and high quality multilevel scheme for partitioning irregular graphs.
\newblock {\em SIAM J. Sci. Comput.}, 20(1):359--392, December 1998.

\bibitem{karypis1998fast}
George Karypis and Vipin Kumar.
\newblock A fast and high quality multilevel scheme for partitioning irregular graphs.
\newblock {\em SIAM Journal on scientific Computing}, 20(1):359--392, 1998.

\bibitem{kazemi2020representation}
Seyed~Mehran Kazemi, Rishab Goel, Kshitij Jain, Ivan Kobyzev, Akshay Sethi, Peter Forsyth, and Pascal Poupart.
\newblock Representation learning for dynamic graphs: A survey, 2020.

\bibitem{khandelwal2017zipg}
Anurag Khandelwal, Zongheng Yang, Evan Ye, Rachit Agarwal, and Ion Stoica.
\newblock Zipg: A memory-efficient graph store for interactive queries.
\newblock In {\em Proceedings of the 2017 ACM International Conference on Management of Data}, pages 1149--1164, 2017.

\bibitem{CuSha2014}
Farzad Khorasani, Keval Vora, Rajiv Gupta, and Laxmi~N. Bhuyan.
\newblock Cusha: Vertex-centric graph processing on gpus.
\newblock In {\em Proceedings of the 23rd International Symposium on High-Performance Parallel and Distributed Computing}, HPDC '14, pages 239--252, New York, NY, USA, 2014. Association for Computing Machinery.

\bibitem{GRIP2020}
Kevin {Kiningham}, Christopher {Re}, and Philip {Levis}.
\newblock {GRIP: A Graph Neural Network Accelerator Architecture}.
\newblock {\em arXiv e-prints}, page arXiv:2007.13828, July 2020.

\bibitem{GCN2017}
Thomas~N. Kipf and Max Welling.
\newblock {Semi-Supervised Classification with Graph Convolutional Networks}.
\newblock In {\em Proceedings of the 5th International Conference on Learning Representations}, ICLR '17, 2017.

\bibitem{graphone19}
Pradeep Kumar and H.~Howie Huang.
\newblock {GraphOne}: A data store for real-time analytics on evolving graphs.
\newblock In {\em 17th USENIX Conference on File and Storage Technologies (FAST 19)}, pages 249--263, Boston, MA, February 2019. USENIX Association.

\bibitem{lamb2016professor}
Alex~M Lamb, Anirudh~Goyal ALIAS PARTH~GOYAL, Ying Zhang, Saizheng Zhang, Aaron~C Courville, and Yoshua Bengio.
\newblock Professor forcing: A new algorithm for training recurrent networks.
\newblock {\em Advances in neural information processing systems}, 29, 2016.

\bibitem{trafficbenchmark}
Fuxian Li, Jie Feng, Huan Yan, Guangyin Jin, Fan Yang, Funing Sun, Depeng Jin, and Yong Li.
\newblock Dynamic graph convolutional recurrent network for traffic prediction: Benchmark and solution.
\newblock {\em ACM Trans. Knowl. Discov. Data}, 17(1), feb 2023.

\bibitem{CBGNN}
Haoyang Li and Lei Chen.
\newblock Cache-based gnn system for dynamic graphs.
\newblock In {\em Proceedings of the 30th ACM International Conference on Information \& Knowledge Management}, CIKM '21, page 937–946, New York, NY, USA, 2021. Association for Computing Machinery.

\bibitem{li2018dcrnntraffic}
Yaguang Li, Rose Yu, Cyrus Shahabi, and Yan Liu.
\newblock Diffusion convolutional recurrent neural network: Data-driven traffic forecasting.
\newblock In {\em International Conference on Learning Representations (ICLR '18)}, 2018.

\bibitem{PaGraph2020}
Zhiqi Lin, Cheng Li, Youshan Miao, Yunxin Liu, and Yinlong Xu.
\newblock Pagraph: Scaling gnn training on large graphs via computation-aware caching.
\newblock In {\em Proceedings of the 11th ACM Symposium on Cloud Computing}, SoCC '20, pages 401--415, New York, NY, USA, 2020. Association for Computing Machinery.

\bibitem{Lo2018}
Yu-Chen Lo, Stefano~E. Rensi, Wen Torng, and Russ~B. Altman.
\newblock Machine learning in chemoinformatics and drug discovery.
\newblock {\em Drug Discovery Today}, 23(8):1538 -- 1546, 2018.

\bibitem{TM-GCN}
Osman~Asif Malik, Shashanka Ubaru, Lior Horesh, Misha~E. Kilmer, and Haim Avron.
\newblock Dynamic graph convolutional networks using the tensor m-product.
\newblock In {\em Proceedings of the 2021 {SIAM} International Conference on Data Mining ({SDM})}, pages 729--737. Society for Industrial and Applied Mathematics, jan 2021.

\bibitem{Manessi2020}
Franco Manessi, Alessandro Rozza, and Mario Manzo.
\newblock Dynamic graph convolutional networks.
\newblock {\em Pattern Recognition}, 97:107000, Jan 2020.

\bibitem{graphbolt19}
Mugilan Mariappan and Keval Vora.
\newblock Graphbolt: Dependency-driven synchronous processing of streaming graphs.
\newblock In {\em Proceedings of the Fourteenth EuroSys Conference 2019}, EuroSys '19, pages 25:1--25:16, New York, NY, USA, 2019. ACM.

\bibitem{Marius2021}
Jason Mohoney, Roger Waleffe, Henry Xu, Theodoros Rekatsinas, and Shivaram Venkataraman.
\newblock Marius: Learning massive graph embeddings on a single machine.
\newblock In {\em 15th {USENIX} Symposium on Operating Systems Design and Implementation ({OSDI} 21)}, pages 533--549. {USENIX} Association, July 2021.

\bibitem{PinnerSage2020}
Aditya Pal, Chantat Eksombatchai, Yitong Zhou, Bo~Zhao, Charles Rosenberg, and Jure Leskovec.
\newblock Pinnersage: Multi-modal user embedding framework for recommendations at pinterest.
\newblock In {\em Proceedings of the 26th ACM SIGKDD International Conference on Knowledge Discovery \& Data Mining}, KDD '20, pages 2311--2320, New York, NY, USA, 2020. Association for Computing Machinery.

\bibitem{gnnforcovid}
George Panagopoulos, Giannis Nikolentzos, and Michalis Vazirgiannis.
\newblock Transfer graph neural networks for pandemic forecasting.
\newblock In {\em Proceedings of the AAAI Conference on Artificial Intelligence}, volume~35, pages 4838--4845, 2021.

\bibitem{pareja2019evolvegcn}
Aldo Pareja, Giacomo Domeniconi, Jie Chen, Tengfei Ma, Toyotaro Suzumura, Hiroki Kanezashi, Tim Kaler, Tao~B. Schardl, and Charles~E. Leiserson.
\newblock Evolvegcn: Evolving graph convolutional networks for dynamic graphs, 2019.

\bibitem{Park2019}
Namyong Park, Andrey Kan, Xin~Luna Dong, Tong Zhao, and Christos Faloutsos.
\newblock Estimating node importance in knowledge graphs using graph neural networks.
\newblock In {\em Proceedings of the 25th ACM SIGKDD International Conference on Knowledge Discovery \& Data Mining}, KDD '19, pages 596--606, New York, NY, USA, 2019. Association for Computing Machinery.

\bibitem{Rossetti2018}
Giulio Rossetti and R\'{e}my Cazabet.
\newblock Community discovery in dynamic networks: A survey.
\newblock {\em ACM Comput. Surv.}, 51(2), feb 2018.

\bibitem{rozemberczki2021pytorch}
Benedek Rozemberczki, Paul Scherer, Yixuan He, George Panagopoulos, Alexander Riedel, Maria Astefanoaei, Oliver Kiss, Ferenc Beres, , Guzman Lopez, Nicolas Collignon, and Rik Sarkar.
\newblock {PyTorch Geometric Temporal: Spatiotemporal Signal Processing with Neural Machine Learning Models}.
\newblock In {\em Proceedings of the 30th ACM International Conference on Information and Knowledge Management}, pages 4564--4573, 2021.

\bibitem{sankar2020dysat}
Aravind Sankar, Yanhong Wu, Liang Gou, Wei Zhang, and Hao Yang.
\newblock Dysat: Deep neural representation learning on dynamic graphs via self-attention networks.
\newblock In {\em Proceedings of the 13th international conference on web search and data mining}, pages 519--527, 2020.

\bibitem{seo2016structured}
Youngjoo Seo, Micha{\"e}l Defferrard, Pierre Vandergheynst, and Xavier Bresson.
\newblock Structured sequence modeling with graph convolutional recurrent networks, 2016.

\bibitem{mo2017gpma}
Mo~Sha, Yuchen Li, Bingsheng He, and Kian-Lee Tan.
\newblock Accelerating dynamic graph analytics on gpus.
\newblock {\em Proc. VLDB Endow.}, 11(1):107–120, September 2017.

\bibitem{DGNNSurvey}
Joakim Skarding, Bogdan Gabrys, and Katarzyna Musial.
\newblock Foundations and modelling of dynamic networks using dynamic graph neural networks: {A} survey.
\newblock {\em CoRR}, abs/2005.07496, 2020.

\bibitem{song2019session}
Weiping Song, Zhiping Xiao, Yifan Wang, Laurent Charlin, Ming Zhang, and Jian Tang.
\newblock Session-based social recommendation via dynamic graph attention networks.
\newblock In {\em Proceedings of the Twelfth ACM international conference on web search and data mining}, pages 555--563, 2019.

\bibitem{RecDGAN2019}
Weiping Song, Zhiping Xiao, Yifan Wang, Laurent Charlin, Ming Zhang, and Jian Tang.
\newblock Session-based social recommendation via dynamic graph attention networks.
\newblock In {\em Proceedings of the Twelfth ACM International Conference on Web Search and Data Mining}, WSDM '19, page 555–563, New York, NY, USA, 2019. Association for Computing Machinery.

\bibitem{Stokes2020}
Jonathan~M. Stokes, Kevin Yang, Kyle Swanson, Wengong Jin, Andres Cubillos-Ruiz, Nina~M. Donghia, Craig~R. MacNair, Shawn French, Lindsey~A. Carfrae, Zohar Bloom-Ackermann, Victoria~M. Tran, Anush Chiappino-Pepe, Ahmed~H. Badran, Ian~W. Andrews, Emma~J. Chory, George~M. Church, Eric~D. Brown, Tommi~S. Jaakkola, Regina Barzilay, and James~J. Collins.
\newblock A deep learning approach to antibiotic discovery.
\newblock {\em Cell}, 180(4):688 -- 702.e13, 2020.

\bibitem{Dorylus2021}
John Thorpe, Yifan Qiao, Jonathan Eyolfson, Shen Teng, Guanzhou Hu, Zhihao Jia, Jinliang Wei, Keval Vora, Ravi Netravali, Miryung Kim, and Guoqing~Harry Xu.
\newblock Dorylus: Affordable, scalable, and accurate {GNN} training with distributed {CPU} servers and serverless threads.
\newblock In {\em 15th USENIX Symposium on Operating Systems Design and Implementation (OSDI 21)}, pages 495--514. USENIX Association, July 2021.

\bibitem{GAT2018}
Petar Veli{\v c}kovi{\'c}, Guillem Cucurull, Arantxa Casanova, Adriana Romero, Pietro Li{\`o}, and Yoshua Bengio.
\newblock Graph attention networks.
\newblock In {\em International Conference on Learning Representations}, 2018.

\bibitem{PiPAD}
Chunyang Wang, Desen Sun, and Yuebin Bai.
\newblock Pipad: Pipelined and parallel dynamic gnn training on gpus.
\newblock In {\em Proceedings of the 28th ACM SIGPLAN Annual Symposium on Principles and Practice of Parallel Programming}, PPoPP '23, page 405–418, New York, NY, USA, 2023. Association for Computing Machinery.

\bibitem{flexgraph21}
Lei Wang, Qiang Yin, Chao Tian, Jianbang Yang, Rong Chen, Wenyuan Yu, Zihang Yao, and Jingren Zhou.
\newblock Flexgraph: A flexible and efficient distributed framework for gnn training.
\newblock In {\em Proceedings of the Sixteenth European Conference on Computer Systems}, EuroSys '21, page 67–82, New York, NY, USA, 2021. Association for Computing Machinery.

\bibitem{DGLPaper2019}
Minjie {Wang}, Da~{Zheng}, Zihao {Ye}, Quan {Gan}, Mufei {Li}, Xiang {Song}, Jinjing {Zhou}, Chao {Ma}, Lingfan {Yu}, Yu~{Gai}, Tianjun {Xiao}, Tong {He}, George {Karypis}, Jinyang {Li}, and Zheng {Zhang}.
\newblock {Deep Graph Library: A Graph-Centric, Highly-Performant Package for Graph Neural Networks}.
\newblock {\em arXiv e-prints}, page arXiv:1909.01315, September 2019.

\bibitem{GNNAdvisor2021}
Yuke Wang, Boyuan Feng, Gushu Li, Shuangchen Li, Lei Deng, Yuan Xie, and Yufei Ding.
\newblock {GNNAdvisor}: An adaptive and efficient runtime system for {GNN} acceleration on {GPUs}.
\newblock In {\em 15th USENIX Symposium on Operating Systems Design and Implementation (OSDI 21)}, pages 515--531. USENIX Association, July 2021.

\bibitem{teacherforcing1989}
Ronald~J. Williams and David Zipser.
\newblock A learning algorithm for continually running fully recurrent neural networks.
\newblock {\em Neural Computation}, 1(2):270--280, 1989.

\bibitem{gpl2021}
Jingqi Wu, Rong Chen, and Yubin Xia.
\newblock {\em Fast and Accurate Optimizer for Query Processing over Knowledge Graphs}, page 503–517.
\newblock Association for Computing Machinery, New York, NY, USA, 2021.

\bibitem{GIN2018}
Keyulu Xu, Weihua Hu, Jure Leskovec, and Stefanie Jegelka.
\newblock How powerful are graph neural networks?
\newblock {\em CoRR}, abs/1810.00826, 2018.

\bibitem{GNNLab22}
Jianbang Yang, Dahai Tang, Xiaoniu Song, Lei Wang, Qiang Yin, Rong Chen, Wenyuan Yu, and Jingren Zhou.
\newblock Gnnlab: a factored system for sample-based gnn training over gpus.
\newblock In {\em Proceedings of the Seventeenth European Conference on Computer Systems}, EuroSys '22, page 417–434, New York, NY, USA, 2022. Association for Computing Machinery.

\bibitem{PinSage2018}
Rex Ying, Ruining He, Kaifeng Chen, Pong Eksombatchai, William~L. Hamilton, and Jure Leskovec.
\newblock Graph convolutional neural networks for web-scale recommender systems.
\newblock In {\em Proceedings of the 24th ACM SIGKDD International Conference on Knowledge Discovery \& Data Mining}, KDD '18, pages 974--983, New York, NY, USA, 2018. Association for Computing Machinery.

\bibitem{STGCN}
Bing Yu, Haoteng Yin, and Zhanxing Zhu.
\newblock Spatio-temporal graph convolutional neural network: {A} deep learning framework for traffic forecasting.
\newblock {\em CoRR}, abs/1709.04875, 2017.

\bibitem{AGL}
Dalong Zhang, Xin Huang, Ziqi Liu, Jun Zhou, Zhiyang Hu, Xianzheng Song, Zhibang Ge, Lin Wang, Zhiqiang Zhang, and Yuan Qi.
\newblock Agl: A scalable system for industrial-purpose graph machine learning.
\newblock {\em Proc. VLDB Endow.}, 13(12):3125–3137, aug 2020.

\bibitem{understandGNN22}
Hengrui Zhang, Zhongming Yu, Guohao Dai, Guyue Huang, Yufei Ding, Yuan Xie, and Yu~Wang.
\newblock Understanding {GNN} computational graph: {A} coordinated computation, io, and memory perspective.
\newblock In Diana Marculescu, Yuejie Chi, and Carole{-}Jean Wu, editors, {\em Proceedings of Machine Learning and Systems 2022, MLSys 2022, Santa Clara, CA, USA, August 29 - September 1, 2022}. mlsys.org, 2022.

\bibitem{zhang2018gaan}
Jiani Zhang, Xingjian Shi, Junyuan Xie, Hao Ma, Irwin King, and Dit-Yan Yeung.
\newblock Gaan: Gated attention networks for learning on large and spatiotemporal graphs.
\newblock {\em arXiv preprint arXiv:1803.07294}, 2018.

\bibitem{zhang2022dynamic}
Mengqi Zhang, Shu Wu, Xueli Yu, Qiang Liu, and Liang Wang.
\newblock Dynamic graph neural networks for sequential recommendation.
\newblock {\em IEEE Transactions on Knowledge and Data Engineering}, 35(5):4741--4753, 2022.

\bibitem{zhang2016exploring}
Mingxing Zhang, Yongwei Wu, Kang Chen, Xuehai Qian, Xue Li, and Weimin Zheng.
\newblock Exploring the hidden dimension in graph processing.
\newblock In {\em OSDI}, volume~16, pages 285--300, 2016.

\bibitem{yu2023egraph}
Yu~Zhang, Yuxuan Liang, Jin Zhao, Fubing Mao, Lin Gu, Xiaofei Liao, Hai Jin, Haikun Liu, Song Guo, Yangqing Zeng, Hang Hu, Chen Li, Ji~Zhang, and Biao Wang.
\newblock Egraph: Efficient concurrent gpu-based dynamic graph processing.
\newblock {\em IEEE Transactions on Knowledge and Data Engineering}, 35(6):5823--5836, 2023.

\bibitem{tgcn2020}
Ling Zhao, Yujiao Song, Chao Zhang, Yu~Liu, Pu~Wang, Tao Lin, Min Deng, and Haifeng Li.
\newblock T-gcn: A temporal graph convolutional network for traffic prediction.
\newblock {\em IEEE Transactions on Intelligent Transportation Systems}, 21(9):3848--3858, Sep 2020.

\bibitem{TGL}
Hongkuan Zhou, Da~Zheng, Israt Nisa, Vasileios Ioannidis, Xiang Song, and George Karypis.
\newblock Tgl: A general framework for temporal gnn training on billion-scale graphs.
\newblock {\em Proc. VLDB Endow.}, 15(8):1572–1580, apr 2022.

\end{thebibliography}

\end{document}